# Combined Federated and Split Learning in Edge Computing for Ubiquitous Intelligence in Internet of Things: State of the Art and Future Directions

Qiang Duan, *Senior Member, IEEE*, Shijing Hu, Ruijun Deng, and Zhihui Lu, *Member, IEEE*

***Abstract*—Federated learning (FL) and split learning (SL) are two emerging collaborative learning methods that may greatly facilitate ubiquitous intelligence in Internet of Things (IoT). Federated learning enables machine learning (ML) models locally trained using private data to be aggregated into a global model. Split learning allows different portions of an ML model to be collaboratively trained on different workers in a learning framework. Federated learning and split learning, each has unique advantages and respective limitations, may complement each other toward ubiquitous intelligence in IoT. Therefore, combination of federated learning and split learning recently became an active research area attracting extensive interest. In this article, we review the latest developments in federated learning and split learning and present a survey on the state-of-the-art technologies for combining these two learning methods in an edge computing-based IoT environment. We also identify some open problems and discuss possible directions for future research in this area with a hope to further arouse the research community's interest in this emerging field.**

## I. Introduction

The recent rapid developments of networking and distributed computing technologies have enabled wide deployment of Internet of Things (IoTs) where a huge amount of data generated from diverse user devices are processed by various applications. Modern IoT applications often leverage machine learning (ML) techniques for provisioning a wide range of intelligent services such as smart-manufacture, smart-transportation, smart-health, etc. The highly disperse nature of big data and widely distributed ML-based applications in IoT calls for ubiquitous intelligence, which focuses on allowing all "things" in IoT, including both user devices and networking/computing equipment, to be involved in the development as well as usage of IoT intelligence. Various IoT devices may be involved in intelligence development through their contributions of data and/or resources for training the ML models that are used by IoT applications.

The huge amount of highly dispersed IoT data make the traditional cloud-based IoT infrastructure insufficient to support applications with strict performance requirements, e.g., real-time operation control in smart-factory and fast incident response in smart-city. The emerging edge computing, which essentially deploys cloud-like capabilities on devices distributed at the network edge, offers a promising infrastructure platform for supporting high-performance IoT applications [1]. Edge computing may facilitate ubiquitous intelligence in IoT by enabling the computation and communication resources on various IoT devices to be fully utilized for training ML models. However, achieving ubiquitous intelligence in an edge computing-based IoT environment still faces some challenges that need to be fully addressed.

A key aspect of ubiquitous intelligence in IoT lies in enabling the highly dispersed big data generated from or collected by the large number of user devices to be fully leveraged for ML model training. Traditional ML techniques assume the entire set of training data be available at a central site like a cloud data center. However, transmitting the huge volume of IoT data to a central data center not only consumes network bandwidth but also introduces extra delay. In addition, the IoT data often contain private information, e.g., the personal information represented by the data collected in smart-health and smart-home scenarios; therefore, in those cases data should be kept on local devices in order to protect user privacy.

Another main challenge to ubiquitous intelligence in IoT is about the resource aspect – how to allow the distributed computational resources in IoT to be fully utilized for ML model training. The edge computing paradigm offers a promising approach to supporting IoT applications by leveraging the distributed resources. However, the edge nodes in an IoT environment are often implemented on IoT devices with constrained computing/networking capabilities that are insufficient for training complex ML models.

The challenges to the data and resource aspects of ubiquitous intelligence in IoT call for new machine learning methods. Collaborative learning is a technical strategy being explored by the research community for addressing the challenges. In general, collaborative learning jointly trains a global model through collaboration without direct access to the decentralized raw data, which are of great appeal to applications for reaping the benefits from the rich data generated in a distributed IoT environment. Federated Learning (FL) and Split Learning (SL) are two representative emerging collaborative learning methods.

Federated Learning (FL) allows multiple data owners to cooperate in training a global ML model without exposing

Qiang Duan is with the Information Sciences & Technology Department, Pennsylvania State University, Abington, PA, 19002 USA (e-mail: qduan@psu.edu).

Shijing Hu and Ruijun Deng are with the School of Computer Science, Fudan University, Shanghai 200082, China

Zhihui Lu is with the School of Computer Science, Fudan University, Shanghai 200082, China and also with the Shanghai Blockchain Engineering Research Center, Shanghai 200000, China (e-mail: lzh@fudan.edu.cn).

their individual datasets [2]. In an FL framework, each data owner (client) trains a local ML model using its own private data set. At the end of each training iteration, the trained models from all clients are aggregated by a server into a global model, which is then used by all clients to start the next iteration of their local training. FL pushes model training to the local devices where user data are stored to avoid transmitting private data from the large number of IoT devices to a central server; therefore, FL offers a promising approach to addressing the challenges to the data aspect of ubiquitous intelligence in IoT. However, FL requires individual user devices to have sufficient resources for training a full ML model, which is not realistic to resource-constrained IoT devices especially for training complex ML models such as deep neural networks.

Split Learning (SL) is another collaborative learning approach in which an ML model is split into two (or multiple) portions that can be trained separately but in collaboration [3]. The user device (client) executes the client-side model using its local data and sends the intermediate results (outputs of the client-side model) to a server. The server uses the received data as inputs to the server-side model to complete the forward propagation. During the back propagation, the gradients are computed at the server and then sent back to the client to complete one training iteration. In a typical SL framework, multiple clients collaborate with one server to train a global model using the private data available on different user devices. SL allows user devices to offload part of the model training task to a server thus makes it possible to leverage the flexible resource management in edge computing for supporting ML model training. Therefore, SL may greatly facilitate the resource aspect of ubiquitous intelligence in IoT. However, the sequential client-server collaboration in SL limits its capability of involving the IoT big data dispersed across a large number of user devices for model training.

Since FL and SL focus on addressing the challenges to ubiquitous intelligence from the data aspect and resource aspect respectively, combining FL and SL in order to exploit the advantages of both learning methods and mitigate their respective drawbacks has appeared as a promising strategy for realizing IoT intelligence. Combined FL and FL recently became an active research topic attracting extensive interest. Although encouraging progress in this area has been reported in the literature, this field is still on its early stage and there lacks a comprehensive survey of the on-going related research. A few survey papers about federated learning have been published, for example [4], [5], [6], and [7]; however, none of these surveys fully cover split learning and its combination with federated learning. In [8], the authors briefly reviewed split learning and some early efforts for combining FL and SL but missed the more recent developments in both SL and hybrid FL-SL techniques.

In this article, we first review the latest developments in federated learning and split learning and then present a survey on the state-of-the-art technologies for combining split learning with federated learning for facilitating ubiquitous intelligence. We also identify some open problems and discuss possible directions for future research in this area with a hope to further arouse the research community's interest in this emerging field. To the best of our knowledge, this is the first survey on split learning and its combination with federated learning in the context of enabling ubiquitous intelligence in an IoT environment.

The rest of this article is organized as follows. In Section II, we give an overview of the FL architecture and representative technologies for FL in IoT. Then in Section III, we introduce the concept of split learning and review the latest development of SL technologies in edge computing-based IoT. A survey on the state-of-the-art technologies for combining split learning with federated learning is presented in Section IV and privacy protection for split learning is discussed in Section V. We identify some open problems for further study and discuss possible directions for future research on SL-FL combination in Section VI. We draw conclusions in Section VII.

## II. Federated Learning in IoT

### A. Introduction to Federated Learning

A general system architecture and the basic working mechanism for federated learning are depicted in Fig. 1. There are two types of entities in the FL system – the data owners that participate in the collaborative model training, which are referred to as FL clients; and the model owner that coordinates the training process and aggregates the models, which is referred to as the FL server. Let $N = \{1, 2, \ldots, K\}$ denote a set of $K$ clients each having a dataset $D_k$ ($k \in K$), then the entire dataset is $D = \cup_{k=1}^{K} D_k$. At the beginning of a learning process, the FL server first initializes a model training task, including specifying the model hyper-parameters and the learning rate, and broadcasts the initial global model to a set of selected clients (step-1 in Fig. 1). Then each selected client $k$ uses its own dataset $D_k$ to train a local model (step-2) and uploads the trained model parameters to the server (step-3). The FL server aggregates the local models received from all the selected clients to generate a global model (step-4) and then updates all clients with the new global model (step-5). The above process of local training–global aggregation is repeated for multiple rounds until a certain level of accuracy is achieved for the global model.

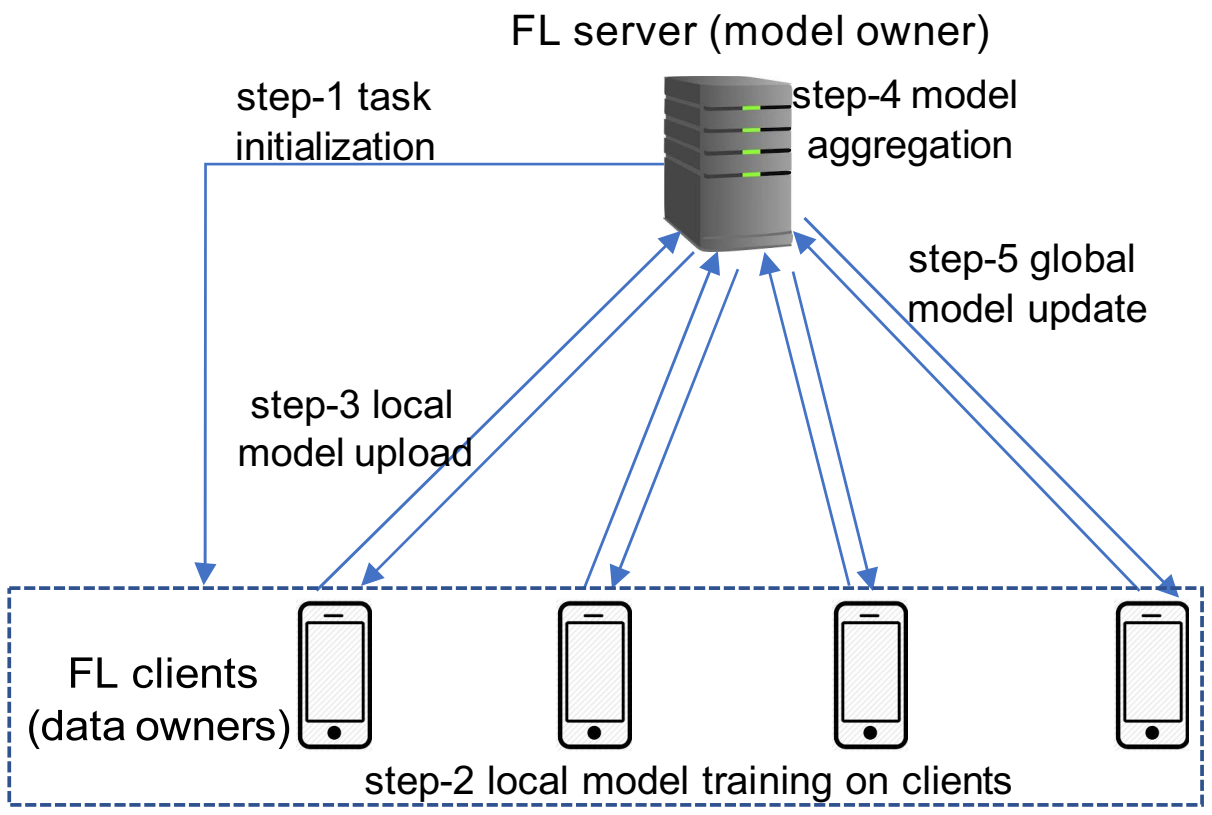

Fig. 1. The framework architecture of federated learning.

The Federated Average (FedAvg) algorithm (given in Algorithm 1) [9] is the most widely accepted algorithm for basic federated learning. The FedAvg algorithm first initializes the global model with parameter $w_0$. Then for each round, the server selects $m$ out of the total $K$ clients for participating in the current round of training and calls the ClientUpdate function for each selected client with the current global parameter $w^G_t$ passed to the client. The ClientUpdate function running on each participating client initializes its local model using the received parameter $w^G_t$ and then trains the local model using the local dataset. After completing $E$ local training epochs, the ClientUpdate function sends the locally trained parameter $w$ back to the server. After collecting local parameters from all the participating clients, the server aggregates all local parameters using weighted average to update the global model. The aggregate weight of each local model is determined by the ratio of the size of the local dataset used for training the model to the size of the entire training dataset.

**Algorithm 1** FedAvg

**Require:** The $K$ clients are indexed by $k$; $B$ is the local minibatch size, $E$ is the number of local epochs, and $\eta$ is the learning rate.

```
Server executes:
initialize w_0
for each round t=1,2,...M do
    S_t ← (random set of m clients)
    for each client k ∈ S_t in parallel do
        w^k_(t+1) ← ClientUpdate (k, w^G_t)
    end for
    w^G_(t+1) ← Σ_{k=1}^{K} (|D_k|/|D|) w^k_(t+1)
end for
ClientUpdate(k, w_t) //Run on client k
initialize w ← w_t
B ← (split P_k into batches of size B)
for each local epoch i from 1 to E do
    for batch b ∈ B do
        w ← w − η ∇ l(w; b)
    end for
end for
return w to server
```

The above FL architecture and mechanism assume a horizontal partition of user data; that is, all the data sets on different clients share the same set of features but contain different data samples. Training data in FL may also be partitioned vertically; that is, all clients share the same set of data samples but differ in feature space. For example, a bank and an e-commerce company in the same city may share a set of customers but have different feature spaces such as credit rating and revenue information at the bank and the browsing and purchasing history at the e-commerce company. Vertical FL has a similar system architecture as shown in Fig. 1 but operates with a different working mechanism. Each client in vertical FL trains a sub-model associated with its own feature space and the model aggregation at the server is often based on sub-model concatenation instead of weighted average as in horizontal FL. We focus on horizontal FL in the rest of this section since the majority of research on FL assumes a horizontal data partition. Readers who are interested in vertical FL are referred to [10], [11], [12], and [13] for more information.

Compared to conventional cloud-centric ML model training, FL has advantages in communication efficiency and data privacy protection. Instead of sending raw data to a data center for training an ML model, FL only transmits model parameters to the FL server for aggregation and update. Therefore, FL may significantly reduce communication cost thus relieving burden of networking, which is particularly beneficial in an IoT environment where a large number of devices are connected via network links with constrained capacities. Since the data used for model training in FL are stored only at the devices where they are generated, privacy of such user data is protected. With enhanced data privacy protection, more data owners are willing to participate in the collaborative model training. This is particularly beneficial to ubiquitous intelligence in IoT which relies on the involvement of a large number of devices each of which contributes a relatively small set of data for training ML models.

### *B. Challenges to Federated Learning in Edge-based IoT*

The main challenges to federated learning in an edge computing-based IoT environment are discussed in this subsection. These challenges mainly come from the special characteristic of IoT, including the highly diverse data generated from a wide variety of user devices, the heterogeneous nodes in the edge computing system with constrained computing and networking capacities, and the dynamic features of a large-scale edge computing-based IoT.

*a)* Statistical Heterogeneity*:* Conventional distributed ML methods often assume independent and identical distribution (iid) for the training datasets. However, the training data used in federated learning are generated on different user devices therefore show heterogeneity in both data volumes and statistical features. This is particularly true for FL in an IoT environment where highly diverse IoT devices, ranging from smart phones and tablet devices generating large volumes of multimedia data to sensors and actuators with small amount of measurement/control data, all participate in federated learning.

*b)* System Heterogeneity*:* An IoT environment comprises heterogeneous computing and networking systems that are implemented with various hardware/software technologies and managed by different providers, for example servers in 5G core network with abundant computational and communication resources and access-points in an ad-hoc sensor network with constrained computing and networking capacities. The wide variety of IoT devices and their network connections provide highly diverse computation and communication capabilities to FL applications, which cause issues such as straggler clients that may dramatically impact FL performance.

*c)* Communication Overheads*:* User devices in IoT are often connected to wireless mobile networks via radio communication links with constrained bandwidth. Although FL reduces communication cost by sharing model parameters rather than raw training data, transmissions of complex ML models (e.g., deep neural networks) from a large number of clients still generate considerable amount of traffic, which makes data communications a potential performance bottleneck for FL in IoT.

*d)* Constrained Computational Resources*:* FL requires the training of an entire ML model on individual client devices. However, user devices in IoT often have limited amounts of computing and storage resources. Therefore, training an entire model especially a complex model such as CNN is a computing-intensive task that consumes large amounts of processing, storage, and power capacities thus is very challenging to many IoT devices with constrained resources.

*e)* System Scalability and Dynamism*:* FL in an IoT environment may involve a big number of devices in a large-scale network in processing massive amount of data dispersed across the network for collaborative model training. An IoT system is also highly dynamic in resource availability due to various reasons such as varying computing capacities on IoT devices and inconstant bandwidth of wireless communication channels. Therefore, FL in edge-based IoT faces the challenges caused by the large-scale dynamic edge computing system.

*f)* Privacy and Security*:* Although FL makes an important step toward protecting data privacy by sharing model parameters instead of raw data, communicating model updates throughout the training process can nonetheless reveal some sensitive information either to the FL server or to a third-party. The local model training and global model aggregation process in FL is also susceptible to a variety of security threats. FL in an edge computing-based IoT environment allows a wide range of compute and network equipment spanning across different trust domains to be involved in the collaborative learning process, which makes the privacy and security issues even more challenging.

### *C. Enabling Federated Learning in IoT*

In this subsection, we review representative technologies proposed for addressing the challenges to FL in edge-based IoT. Since the focus of this article is split learning and its combination with federated learning, we keep our review of FL technologies concise with an objective of providing the necessary background information for the rest of the article. A more thorough survey on federated learning in edge computing can be found in [5].

*1) Enhancing FL Algorithms and Model Aggregation:* Enhancing FL algorithms and model aggregation mechanisms is one of the main strategies for addressing the challenges of heterogeneity in data distributions and system capabilities to FL in IoT. The technical solutions developed following this strategy roughly belong to two categories: i) enhancing FL algorithms by modifying various aspects of the FedAvg algorithm or employing personalized learning methods in FL, and ii) enhancing model aggregation by controlling when and how local models are aggregated into a global model.

Representative methods for enhancing FL algorithms for addressing statistical/system heterogeneity can be categorized into the following two groups – modifying the learning algorithm and personalizing FL training. Various aspects of the FedAvg algorithm, including the objective function, weights for gradient average, learning rate, etc. may be modified in order to enhance FL performance over the heterogeneous datasets and/or systems. For example, the FedProx algorithm proposed in [14] includes the global model parameters in the optimization objective for local model training on each client to address the statistical heterogeneity. Li et al. [15] proposed a q-Fair FL algorithm that revises the loss function of FedAvg to maintain fairness among heterogeneous clients. Personalized FL assigns different task models to different clients to exploit the correlation among the different tasks in order to capture the relationship among the non-iid datasets across clients. For example, the systems-aware FL method MOCHA proposed in [16] takes into account client capabilities (storage, computation, and communication) for implementing multi-task FL.

A regular FL framework performs synchronous model aggregation; that is, the server updates the global model once after all the participating clients upload their locally trained models. The diverse computing and networking capacities across clients and the different sizes of client datasets cause the model training on some clients (stragglers) much slower than those of others. Therefore, synchronous model aggregation is inefficient for FL in edge-based IoT and may degrade learning performance. Various research efforts have been made to develop new model aggregation mechanisms for FL to address this problem. Representative works can be classified into three categories: asynchronous aggregation, semi-asynchronous aggregation, and semi-synchronous aggregation.

Asynchronous aggregation allows the server to perform aggregation whenever a local model is uploaded from a client. For example, in the FedAsync algorithm proposed by Xie *et al.* [17] the FL server updates the global model using weighted averaging whenever it receives a local model from a client. Semi-asynchronous aggregation is to perform model aggregation when the server receives local model updates from a subset of the client population. In the FedBuff algorithm proposed in [18], the FL server starts model aggregation after receiving updates from $k$ clients in each round of training iteration. The CE-AFL mechanism proposed in [19] aggregates the local models from a certain fraction of all clients in each round of training. A semi-synchronous FL (SSFL) algorithm was proposed in [20], in which the server performs synchronous model aggregation while the clients continuously train on their local datasets up to the synchronization point when all the local models are aggregated.

*2) Client Selection in FL:* Enabling ubiquitous intelligence in IoT via FL not only requires incentive for user devices to participate in collaborative learning but also needs to ensure the participants to make positive contributions to the learning process. The large number of user devices with heterogeneous implementations and diverse datasets make it critical to select an appropriate set of clients in each training round in order to achieve high-performance FL. Therefore, client selection is an important aspect of FL in IoT for addressing the challenges of heterogeneity in system capabilities and data distributions. Typical approaches to client selection in FL use either system capabilities (for computation and communication) or dataset quality or their combination as the selection criteria.

A typical system capability-based method for FL client selection is to use the system status information obtained from the clients for making selection decision. For example, the FedCS framework proposed in [21] allows the server to select the maximum possible number of clients that can complete their local training tasks within a pre-specified deadline using a greedy algorithm that favors clients with larger computing capacities and more bandwidth. However, the large-scale dynamic edge computing in IoT makes timely collection of precise state information from client devices challenging and expensive. Online learning techniques (e.g., reinforcement learning) have been explored to address this issue by enabling client selection without prior knowledge of the environment. For example, the experience-driven control scheme proposed in [22] employs the deep Q-learning technique to adaptively choose client devices with an objective of involving as many clients as possible to minimize the global model training time and energy consumption on client devices.

The significant impact of data heterogeneity on FL performance makes the quality of client datasets an important factor for client selection. For example, the authors of [23] proposed a client selection strategy that in each training round chooses a set of clients whose data generate the largest local loss values with respective to the current global model, which is verified by experiments that achieves faster convergence compared to unbiased client selection. The CSFedAvg FL framework proposed in [24] is another data quality-based client selection method, which aims to choose the clients whose local data distributions are similar to the distribution of the entire training dataset in order to alleviate the degradation in the global model accuracy caused by non-iid data across clients.

FL client selection technologies that jointly use system capabilities and dataset quality as the selection criteria have been developed in order to address the heterogeneity in both system capabilities and data distributions. For example, the client selection method proposed in [25] utilizes the importance sampling technique to jointly address the heterogeneity in computation capacities, communication bandwidth, and dataset distributions in FL selection. In [26], the authors formulate the joint resource allocation and client selection for FL as an optimization problem with an objective of minimizing the training loss while meeting the delay and energy consumption requirements. The computing power, network bandwidth, and dataset quality of the client devices are all considered in this client selection scheme.

*3) Communication Efficient FL:* Communication-efficient FL is an active research area in which various technologies have been developed to overcome the challenge of communication overheads in a large-scale IoT environment. Representative methods for communication-efficient FL can be classified into two main categories: i) reducing the frequency of communications between the clients and server, and ii) reducing the amount of data transmission for each client-server communication session.

A common approach to reducing the frequency of client-server communications is to let the clients to conduct local training as much as possible before model aggregation is needed. For example, Yao *et al.* applied the two-stream model commonly used in transfer learning to FL [27]. In each round of local training, the global model received by the client is used as a reference so that the client learns not only from its own local dataset but also from other participating clients with reference to the fixed global model. Another approach to reducing the frequency of client-server communications is selective model updates – skipping model updates (and the associated client-server communications) for some clients in each training iteration. The CEFL framework proposed in [28] follows this idea and selects local models for uploading based on the similarity between the local and global models.

A trade-off between the computation cost for local training and the communication cost for model updating must be carefully considered in an FL framework order to handle the limitation of both networking and computing resources in IoT. One work toward this direction is the control algorithm proposed in [29] that considers the heterogeneous computing and communication conditions across edge devices. This algorithm attempts to minimize the total costs of computation and communication in FL (the total energy consumption for communication and computation in particular) by controlling the frequency of model aggregation.

Another strategy for communication-efficient FL is to reduce the amount of traffic for each round of client-server communication. Typical methods for reducing FL communication traffic include parameter quantization, model sparsification, and model dropout.

Parameter quantization in FL is to compress the local model parameters through quantizing each parameter to a low precision value thus reducing the total number of bits transmitted to the FL server. For example, the cpSGD framework proposed in [30] adopts a quantization method whereby clients quantize their locally computed gradients and send an efficient representation of the quantized gradients to the server. The scheme proposed in [31] further reduces client-server communication traffic by allowing the clients to send the quantized difference between their current local models and the most recent global model to the server.

Model sparsification in FL is to filter out some of the local model parameters and select only a subset of parameters to be communicated with the server. Compared to quantization, sparsification is believed to be more effective for communication-efficient FL. The top-$k$ sparsification method has been employed in FL to upload only the $k$ parameters with the largest absolute gradient values in a local model, which has been extended to reduce traffic for both up-stream (clients to server) and down-stream (server to clients) communications in FL [32]. Choosing an appropriate sparsification degree (i.e., the $k$ value for top-$k$ sparsification) is critical to achieve an optimal trade-off between communication efficiency and learning performance. An adaptive method FAB-top-$k$ proposed in [33] employs an online learning algorithm for automatically determining and dynamically adjusting the optimal value of $k$ for performing top-$k$ sparsification for both upstream and downstream communications.

Another type of sparsification method is to select parameters for communication based on their importance to contributing the global model training. One method is to evaluate parameter importance based on their impact on the loss function. For example, the eSGD algorithm proposed in [34] keeps track of the loss function values at two consecutive training iterations and sends the current gradients to the server only when the loss value of the current iteration decreases comparing to the preceding iteration (which implies that the current gradients are important to the learning process).

Unlike the sparsification methods that drop out some of the parameters after the local model has been trained in each round, the idea of the model dropout approach is to let each client only trains a sub-model – a "sparsized" global model with a subset of parameters dropped out. Model dropout in FL not only improves communication efficiency by exchanging sub-models between clients and server but also reduces the computing loads on clients by training only a sub-model. For example, In the federated dropout method proposed in [35], the server drops out a fixed percentage of parameters from the global model to form the sub-model sent to clients. Each client trains the sub-model it received from the server and sends its update back to the server, which then maps sub-model updates back to the global model.

*4) Privacy and Security Protection in FL:* Privacy and security protection is critical to FL especially when FL frameworks are deployed upon an edge computing platform in an IoT environment. Privacy threats to FL mainly aim at inference of training samples/labels and properties of the training data such as membership and class representatives. The representative approaches to privacy protection in FL include differential privacy-based approaches and encryption-based approaches.

Differential privacy (DP) is the most widely used approach to FL privacy protection mainly due to its strong information theoretic basis and algorithmic simplicity [36]. A common approach to employing DP in FL is to add some "noise" to the trained parameters using a differential privacy randomized mechanism (e.g., a Gaussian mechanism or Laplace mechanism) before sending the parameters to the server. Examples of DP-based privacy protection in FL include the frameworks proposed in [37] and [36]. DP may also be achieved via multiplicative perturbation methods that transform the original data into another space, for example the federated multiplicative update (FMU) algorithm developed in [38] and the matrix multiplicative perturbation method employed in [39].

The main encryption-based techniques for privacy protection in FL include homomorphic encryption (HE) and secure multiparty computation (SMC). HE is a type of encryption with ability to perform computation functions directly over encrypted data while achieving the same (encrypted) results as if the functions were run on plain text [40]. HE can be employed in FL to encrypt the local models on clients before uploading them to the server. For example, the HE scheme was used in a three-party FL framework in [41] and a multiparty privacy preserving FL framework in [42]. SMC can be applied in FL to allow the server to compute the aggregation function (e.g., weighted average) of model updates coming from a large number of clients without learning each client's individual contribution. For example, an SMC-based protocol for privacy preserving model aggregation in FL was proposed in [43].

Security attacks to FL aim at compromising the integrity of the learning process by either degrading the model training performance to an unacceptable level or injecting predefined malicious training samples (backdoors) into a victim model while maintaining the performance of the primary task. Based on the employed technical strategies typical security attacks to FL can be classified into two categories – data poisoning attacks and model poisoning attacks. Data poisoning attacks aim to inject malicious data into the training dataset before the learning process starts while model poisoning attacks attempt to directly poison the local models by either modifying the original ones or forging new ones in order to sway the global model learning toward some malicious objectives.

Typical technologies for defending FL security include malicious participant detection and malicious impact mitigation. Malicious detection methods aim to identify events that do not conform to the expected behavioral or statistical patterns in normal FL training process thus indicating malicious participants. Then various respond actions may be taken to the detected malicious participants, ranging from removing it completely from the training process to reducing its weight in model aggregation. For example, the FL frameworks proposed in [44], [45], and [46] employ this type of security technologies. Representative methods for mitigating malicious impact for defending FL security include majority vote as employed in [47] and normalization as proposed in [48].

## III. Split Learning in IoT

### A. Split Learning Framework Architecture

The general architecture of an SL framework is illustrated in Fig. 2. In the framework, an ML model (usually a neural network) is split into two portions – the client-side model $W_C$ and the server-side model $W_S$ that are executed respectively at a client and a server. Similar to FL, all the raw training data are stored on the client without being transmitted to the server. The training of the full model is done by executing a sequence of forward/backward propagation between the client and server. The client uses the training data to feed the model $W_C$ and performs forward propagation until the *cut layer* (the last layer of $W_C$). Then the cut layer's activations, called *smashed data*, are transmitted (typically together with the corresponding label) to the server. The server uses the smashed data received from the client as the inputs to its model $W_S$ and completes forward propagation on the remainder of the full model. After calculating the loss function, the server starts the back propagation process in which it computes gradients and updates the weights of each layer of $W_S$ until reaching the cut layer. Then the server transmits the gradients of smashed data back to the client. Upon receiving the gradients from the server, the client executes its back propagation on $W_C$ to complete a single pass of back propagation of the full model. In SL, the forward/back propagation between the client and server continues until a convergence point is reached for the full model.

An SL framework with multiple clients uses data from multiple entities in a round robin fashion – all clients take turns with alternating epochs in working with the server.

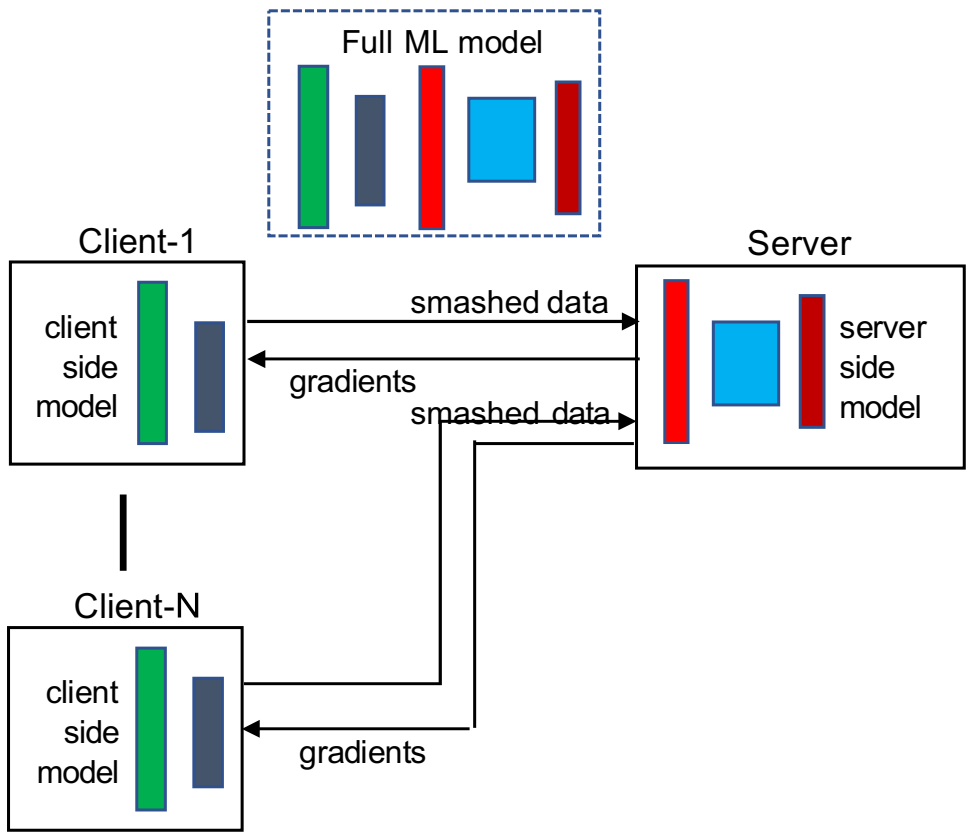

Fig. 2. Framework architecture for multi-client split learning.

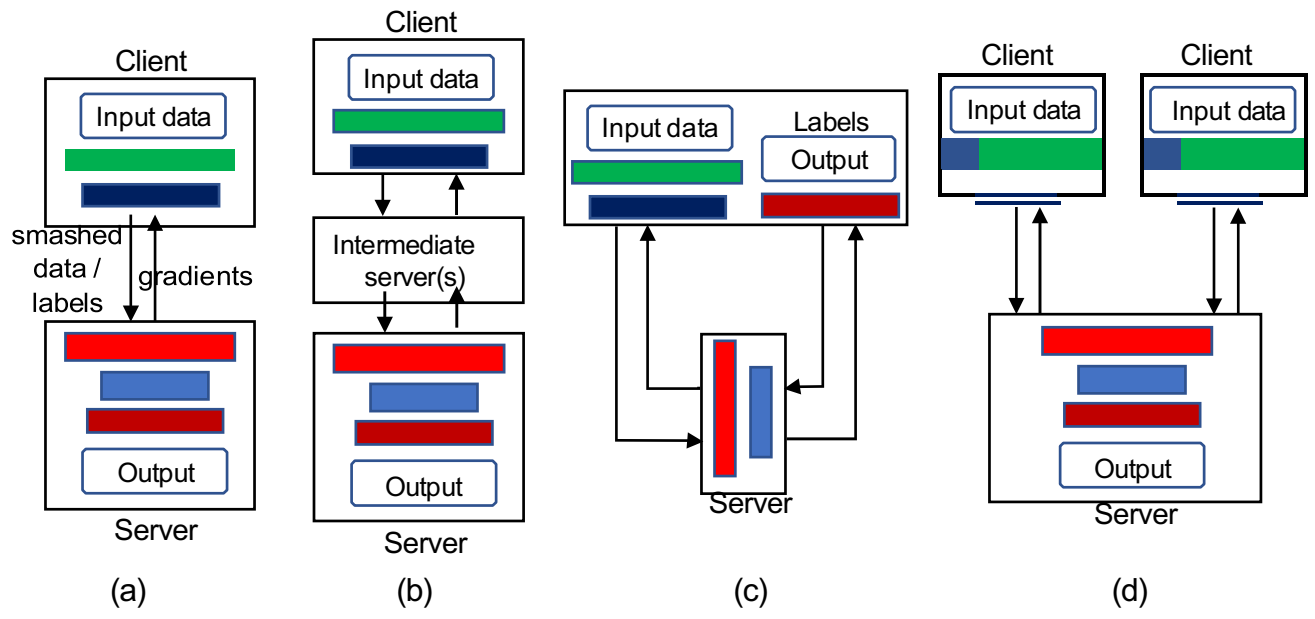

Fig. 3. Configurations of split learning frameworks

Synchronization of the sub-models on different clients is needed for model consistency in multi-client SL; therefore, each client is required to update its model weights before starting the next training epoch. Client-side models can be synchronized in two modes – centralized synchronization and peer-to-peer synchronization. In the centralized mode, a client uploads the weights of its just trained model to the server (or a third-party server) from where the next client in the training order downloads these weights. In the peer-to-peer mode, the server sends the last trained client's address to the current training client, who will use the address to connect the last client and download the model weights. It is possible for the client-side models in a multi-client SL framework to be trained without synchronization; however, there is no convergence guarantee for such asynchronous model training.

There are multiple configurations for an SL framework as shown in Fig. 3 [8]. The basic configuration is often referred to as "vanilla" SL and has been described as the typical SL architecture, in which the ML model is split to two portions for a pair of client and server that share the smashed data and labels (Fig. 3(a)). In an extended SL configuration, the ML model is split to multiple portions and some other workers process some intermediate layers of the model before passing it to the main server (Fig. 3(b)). A multi-hop SL framework is a special type of extended SL configuration in which one client has data and multiple servers in a sequence each trains a portion of the ML model.

In the configuration for SL without label sharing (Fig. 3(c)), the client only transmits the smashed data (not the labels) to the server, which completes forward propagation up to some layers of the model called the *server cut layer* and then sends the activations of the server cut layer back to the client. Then the client completes the forward propagation up to the last layer of the full model. Afterward, the client starts back propagation and sends the gradients of the server cut layer activations to the server. Then the server carries out the back propagation and sends the smashed data's gradients to the client, which then completes the entire back propagation process. Since the process flow of the forward/back propagation in this configuration forms a U shape (from the client to the server and then back to the client for the forward propagation), this configuration is also referred to as "U-Shape" SL.

In the vertical SL configuration, both the dataset and the client-side model is partitioned vertically (Fig. 3(d)). That is, each client owns a subset of features of the sample data and uses them to train a portion of the client-side model that processes the corresponding features. Then the clients transfer their smashed data to the server, which concatenates the smashed data from all clients and carries on the forward propagation on the single server-side model. The back propagation proceeds from the output layer up to the concatenation layer at the server, which then transmits the respective smashed data's gradients to the clients. Then, the clients perform back propagation on their portions of the client-side model.

### B. *Split Learning Performance*

The performance of SL as an emerging collaborative learning method still needs to be thoroughly studied. The two main aspects of SL performance are i) learning performance including convergence (if the trained model converges), convergence rate (how fast the model converges), and model accuracy (how well the model converges); and ii) training costs including computation and communication overheads. Since SL is often regarded as an alternative to or an extension of FL, comparison between the performance of SL and FL in various IoT settings will be insightful to choose the most appropriate learning method for a given IoT setting.

Empirical evaluations of SL learning performance and training costs were presented in [49] together with comparison to FL performance. Multiple data sets with various statistical features, including iid and balanced data, iid and imbalanced data, and non-iid data, were used in the experiments for evaluating SL learning performance. The obtained experiment results indicate that SL may achieves a higher convergence rate than FL with balanced and imbalanced iid data distributions. However, SL learning performance appears to be very sensitive to various settings including data distribution and system scale (the total number of clients). SL converges slower than FL with non-iid data or even fails to converge under some extreme non-iid data distributions.

For evaluating SL training costs, the authors of [49] deployed both FL and SL frameworks on Raspberry Pi devices and compared the associated communication and computation overheads. It was found that FL appears to perform better over SL in an IoT setting where the communication traffic is the primary concern because FL generates a lower amount of communication overheads compared with SL. Such an observation indicates that SL is recommended to achieve better model accuracy and faster convergence in an IoT environment where communication is not a significant concern, e.g., having high-throughput connections provided by an Ethernet or 5G network. However, FL is preferred over SL in application scenarios where communication may become a bottleneck.

A basic idea behind SL lies in the trade-off between computation and communication; that is, SL aims to reduce computation overheads on client devices by splitting the model training to multiple collaborating workers, which may introduce additional communication overheads due to the frequent interactions among the workers. Therefore, communication efficiency is an important aspect of SL performance.

An analysis on SL communication efficiency was reported in [50]. Suppose there are $k$ clients involved in training a full model with $N$ parameters using a dataset with total size $p$, if the SL splits a fraction $\eta$ of the model to the client-side, then according to the analysis in [50] the communication overheads for each client and for the entire SL system in each epoch are given in Table I. The analysis assumes that each client has a equal share of the entire dataset, i.e., $p/K$ data samples per client. Then the communication costs for both forward and back propagation will be $(p/K)q$ (for transmitting activations and gradients respectively). The analysis considered two cases – SL with and without the client-side model synchronization that adds a communication cost $\eta N$ for transmitting the client-side model parameters. Table I also includes FL communication overheads for comparison.

The main impact factors of SL communication overheads include the size of the cut layer $q$ that determines the amount of smashed data transmitted between the client and server per iteration, the dataset size $p$ that determines the number of iterations in each epoch, the total number of clients $K$, and the (client-side) model size $\eta N$ that effects the communication overheads for model synchronization. The obtained analysis results indicate that SL is more communication efficient compared to FL in an IoT environment where a large number of clients collaborate for training a large ML model. On the other hand, FL is a better option in terms of communication overheads in an IoT setting with a large volume of training data but a moderate number of clients and small model size.

The learning performance of vertical SL was evaluated in [51], where model accuracy and F1-score were used as the performance metrics and three financial datasets were used in the experiments. Due to the small number of features of the training data samples, two of the three datasets were split in half and the third dataset was partitioned to four portions. It is worth mentioning that from a practical standpoint the number of clients in vertical SL is likely to be small since data are partitioned according to features. The obtained experiment results verify that vertical SL may achieve the same level of learning performance as that of single (non-split) model training. Experiment results also show that the communication overheads of vertical SL is dependent on the size of the output at the endpoints layer while the computation cost is dependent on the architecture and the size of the input feature vector at each layer.

The aforementioned works on SL performance typically assume deep neural network models such as 2D CNN. However, sequential/time-series data are also common in an IoT environment. In [52], the authors evaluated the performance of SL when this emerging learning method is applied to the 1D CNN model that is trained using sequential data. The authors applied SL to two recent 1D CNN models given in [53] and [54] that are shown the best-achieved accuracy. The model accuracy of the split version of these two 1D CNN models were tested using MIT-BIH arrhythmia database, which is a popular dataset for ECG signal classification or arrhythmia diagnosis detection models. The obtained results show that the split 1D CNN model is able to achieve the same level of accuracy as the non-split 1D CNN, which indicates that split learning is applicable to 1D CNN without performance degradation.

### C. *Enhancing Split Learning in IoT*

The flexibility of training portions of a large ML model on different devices introduced by SL may significantly facilitate ubiquitous intelligence in IoT with constrained computational resources. However, the advantage of SL is often achieved at a cost of increase in bandwidth consumption and may result in sub-optimal convergence especially with heterogeneous client data. In addition, multiple clients in SL take turns to train a shared server-side model sequentially thus may prolong the training time. Some research efforts have been made for addressing these issues to enhance SL performance in an IoT environment.

A typical strategy for improving communication efficiency in SL is to reduce the frequency of communications between the clients and the server. The main cause for the increased communication overheads in SL lies in the dependence of clients on the server for obtaining training gradients, which makes each client to interact with the server in every training iteration (vs. once-per-training round in FL). Therefore, researchers have explored decoupling the clients and server in the SL framework through some kind of asynchronous training schemes that may achieve acceptable learning performance with less frequent data transmissions for exchanging activations and gradients.

A loss-based asynchronous training scheme was proposed in [55] for reducing SL communication overheads. In this scheme, the server-side model is trained as usual while the client-side model is only updated when the loss difference with that of the last update is larger than a pre-defined threshold. The SL framework calculates a loss function on the sever and updates a *state* based on the loss function value in each epoch, following the state diagram shown in Fig. 4. When *state* = A, both activation and gradient are transferred between clients and server. When *state* = B, the activation is sent to the server but no gradient is sent back to the client. When *state* = C, there is no client-server communication, and the server uses the previous activation of cut layer for training the server-side model. The client-side model is not updated when *state* = B or C, which also reduces computation on the client device.

TABLE I
COMPARISON OF SL AND FL COMMUNICATION OVERHEADS.

| Methods | Communication cost per client | Total communication cost |
|---|---|---|
| SL with synchronization | $(p/K)q + (p/K)q + \eta N$ | $2pq + \eta NK$ |
| SL without synchronization | $(p/K)q + (p/K)q$ | $2pq$ |
| Federated learning | $2N$ | $2KN$ |

Experiment results reported in [55] indicate that significant reduction in communication overheads can be achieved by the proposed scheme with only slight loss in model accuracy.

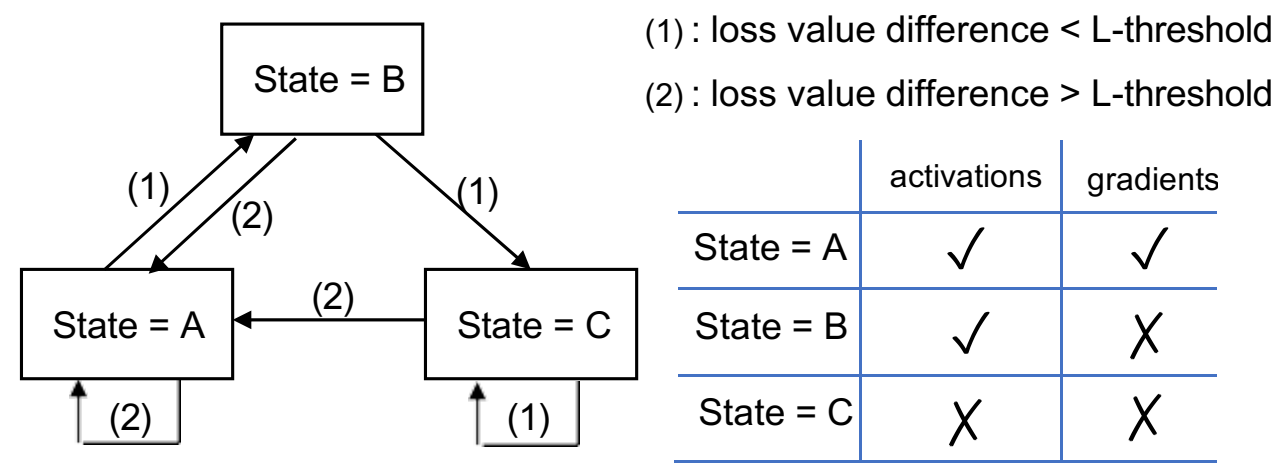

Fig. 4. State diagram for controlling client-server communications in [55].

For reducing SL bandwidth consumption, the AdaSplit framework proposed in [56] divides the total R rounds of training into two phases – the *local phase* and *global phase*. In the *local phase*, training only happens on the client side and each client trains its client-side model using a local loss function; therefore, no data transmission occurs between the client and server. In the *global phase*, activation data are transmitted to the server for training the server-side model but no gradient is sent back to clients. That is, the server-side model is trained using activation received from the clients while the client-side models are still being trained using the local loss functions since no gradient is sent back from the server. In order to make SL scalable to heterogeneous clients, the AdaSplit framework uses an orchestrator during the *global phase* to choose a subset of clients to communicate with the server in each iteration. The orchestrator uses a running statistic of local client losses to select clients and uses a UCB strategy [57] to prioritize clients who need the performance of the server-side model to be improved on their data while ensuring that the full model can be generalized well to the heterogeneous client data distributions.

In the SL framework, multiple clients sequentially update the shared parameters on the server, which often results in the global model converging to sub-optimal accuracy especially with heterogeneous data distributions across clients. One approach to alleviating this issue is to take advantage of the fact neural network models are vastly over-parameterized [58] and only a small proportion of the parameters can learn each client's task with little loss in performance [59]. Following this approach, the AdaSplit framework allows each client to update only a portion of the server-side model to mitigate the impact of heterogeneous client data on model accuracy.

Another approach to addressing the issue caused by sequential client training in SL is to enable parallel training in the multi-client SL framework. Toward this direction, the parallel SL framework proposed in [60] enables parallel training on clients by setting the mini-batch size on each client in proportional to the dataset size on the client so that all clients have the same number of iterations in each epoch. The proposed parallel framework allows all the clients to perform the training of a single mini-batch in parallel and send their activations to the server. The server uses a mini-batch size that is the sum of the mini-batch sizes of all clients; that is, the server uses the activations from all clients generated in one mini-batch to calculate the loss function and gradients once.

## IV. COMBING SPLIT LEARNING AND FEDERATED LEARNING

FL and SL are two paradigms of collaborative learning that may respectively facilitate the data aspect and resource aspect of ubiquitous intelligence. On the other hand, training full models on individual devices in FL limits its capability of utilizing the highly diverse and heterogeneous computational resources in IoT while sequential client-server collaboration in SL restricts its effectiveness in leveraging the big data dispersed across a huge number of IoT devices. Therefore, combination of FL and SL may fully unleash the advantages of both learning methods while mitigating their respective drawbacks. In this section, we review the state-of-the-art technologies for combining FL and SL to present a big picture about the latest developments in this area. The majority of the research on combining FL and SL assumes a horizontal partition of user data with a few recent works considering other configurations (such as vertical and sequential configurations) that are reviewed at the end of this section.

### *A. Hybrid Frameworks for Combining Split Learning with Federated Learning*

Most of the currently available research on combining FL and SL has been conducted from a perspective of enhancing SL performance by applying the FL mechanism. The sequential relay-based training across multiple clients in SL slows down the global model convergence thus forming a bottleneck that limits the scalability of multi-client FL frameworks. In addition, the sequential training in SL may cause the "catastrophic forgetting" issue – the model favors the client data it recently used for training and is prone to forget what it learns from the previous client's data [61]. These issues of SL become severe in an IoT environment where a large number of clients have heterogeneous data distributions. In FL, all clients perform model training in parallel and the server aggregates the trained local models into a global model. Therefore, FL removes the training bottleneck caused by the shared server-side model in SL thus may speedup the training process. Also, FL ensures each client's data to contribute the global model through the model aggregation therefore resolves the "catastrophic forgetting" problem of SL.

A main approach to combining SL with FL is to enable training parallelization and model aggregation in the SL framework. Since in SL the model training is split to two portions that are respectively executed on the client and server sides, training parallelization and model aggregation may be introduced separately on each side. Also, parallel training for server-side models may be deployed on a single central server or distributed to multiple servers.

The SplitFed framework proposed in [62] is an attempt to amalgamate FL and SL by enabling parallel training and aggregation for both client-side and server-side models. In the framework, all clients perform forward propagation on their client-side models in parallel and pass their smashed data to the main server. The main server processes the forward propagation and back-propagation for the server-side model with each client's smashed data separately in parallel. Then the server sends the gradients of the smashed data to the respective clients for back-propagation. After receiving the gradients of its smashed data, each client performs the back-propagation to update its client-side model parameters. At the end of each epoch, the server updates its model using the FedAvg aggregation, i.e., weighted average of the gradients that it computed during the back propagation on each client's smashed data. Similarly, the sub-models trained on all the participating clients will also be aggregated, which can be done by a dedicated server (Fed server) or by the main server (acting as a Fed server). The experiment results reported in [62] for comparative performance evaluation of the proposed framework against regular SL indicate that SplitFed is faster than SL in model convergence with a similar level of model accuracy as SL.

The SplitFed framework requires the server to have sufficient computing power for training multiple instances of the server-side model, one for each participating client, in parallel. Such a requirement becomes overwhelming with the increasing number of clients especially when a large portion of a complex model is split to the server-side. Considering the constrained computational resources available on typical edge nodes in IoT, the server in SplitFed might have to be hosted in a cloud data center; however, the communications between client devices and the remote cloud server may become a bottleneck that deteriorates the system performance.

The authors of [62] also proposed a variant of the SplitFed framework (referred to as SplitFedv2) that enables parallel training and aggregation only for the client-side. In SplitFedv2, the client-side operation remains the same as in SplitFed. On the server-side, the forward-backward propagation is performed sequentially with respect to the clients' smashed data (that is, no aggregation for the server-side model). The server receives the smashed data from all participating clients synchronously and the order in which the corresponding server-side operations are performed for clients is chosen randomly. Compared to SplitFed, SplitFedv2 keeps the shared server-side model training feature of SL that may achieve higher model accuracy. The random order of client-side model training in SplitFedv2 also mitigates the catastrophic forgetting problem caused by the sequential training of basic SL.

A root reason for catastrophic forgetting in SL lies in the default alternative client training order. That is, in each epoch a client completes its training of the client-side model with the shared server-side model using its entire dataset before the client next in order starts training. The SplitFedv3 framework proposed in [63] attempts to address the catastrophic forgetting issue by using alternative mini-batch training (instead of the regular alternative client training). In SplitFedv3, a client trains its model using one mini-batch data samples and then updates the shared server-side model, after which the client next in order takes over. The advantage of alternative mini-batch training over alternative client training is that it avoids sequential training over the entire client dataset for the model, rendering the model training in a more randomized manner thus mitigating the catastrophic forgetting issue over the learning process.

The Cluster-based Parallel SL (CPSL) framework proposed in [64] enables federated parallel training in SL following an idea of "first-parallel-then-sequential." In CPSL, client devices are grouped into multiple clusters and each training round is divided to two stages – parallel intra-cluster training first and then sequential inter-cluster training. All clients in the same cluster perform parallel training in collaboration with the server in a way that is similar to SplitFedv2. After a single round of intra-cluster training is completed, all the client-side models in the cluster are uploaded to the server for aggregation and update. Then the updated model will be used to initialize the client model in the next cluster to start intra-cluster training in that cluster. That is, inter-cluster training in CPSL works in the same way as in the basic multi-client SL framework except that each "client" now consists of a cluster of user devices.

In order to face the challenges to collaborative learning brought in by the highly diverse IoT devices with heterogeneous resources and data distributions, the Hybrid Split and Federated Learning (HSFL) scheme proposed in [65] combines these two learning methods in the same framework from a perspective of client selection and scheduling. The HSFL framework organizes the participating clients into two groups – one group performs federated learning and the other group performs split learning. The server in HSFL chooses the clients for each group in each training round based on the current computing and networking status of clients with an objective of minimizing the total energy consumption on user devices while achieving satisfactory learning performance. The HSFL scheme offers the flexibility to handle dynamic and diverse system environment in which a hybrid SL-FL framework is deployed. On the other hand, the control functions for client selection and scheduling need to be performed by the server for each round of training thus introducing extra complexity and additional computation overheads on the server.

The aforementioned hybrid SL-FL frameworks (SplitFed variants, CPSL, and HSFL) all have a centralized server architecture that deploys the full server-side model training (either parallel or sequential) on a single edge node, which is not scalable with the increasing number of clients. The parallelization of server-side model training may be realized using a decentralized architecture with multiple servers hosted on different edge nodes, which avoids the potential bottleneck that may be caused by a single server performing the parallel training of multiple server-side models.

In the Federated SL (FedSL) framework proposed in [66], there are the same number of servers and clients. Instead of training all server-side models on a single server, the framework trains the server-side model corresponding to each client on an individual server. That is, all client-server pairs perform training of their respective client- and server-side models in parallel. When all data in each client have been used to update the model parameters (i.e., at the end of each training epoch), the server-side models are aggregated by a Fed server and updated on each server. In this framework, each server only works with a single client thus decoupling the performance dependency on other clients. Moreover, the communication pattern in this framework is changed to point-to-point data transfer between each pair of client and server, which avoids potential network congestion caused by data transmissions from multiple user devices to a single edge node as in frameworks with a central server.

In [67], the authors conducted comparative performance evaluation of the FedSL framework against the parallel SL framework proposed in [60]. The obtained experiment results show that the training time of the FedSL framework is constantly shorter that of parallel SL, which indicates that using multiple servers can better unleash the benefit of parallel training for both client and server sides. The FedSL architecture can be especially advantageous as the number of clients increases. It was also found that FedSL achieves a similar level of model accuracy as parallel SL if each client has enough data, but parallel SL converges to a better model if the dataset size at each client is small.

Although the one-to-one client-server pairing scheme proposed in [66] enables federated parallel training in an SL framework, it limits the flexibility of deploying the training jobs on servers. In a typical edge computing-based IoT, the various edge nodes that may host SL servers have highly diverse amounts of computation and communication resources; thus demanding flexible resource allocation for SL server deployment. Toward this direction, a generalized SplitFed learning framework (SFLG) was proposed in [68] to enable a varying number of server-side models that can be deployed on different edge nodes; therefore, one can flexibly choose the number of edge nodes for hosting the servers based on the available edge computing resources.

The architecture of SFLG framework is illustrated in Fig. 5. In SFLG, the clients are divided into multiple groups and each group of clients share one server-side model. The SFLG framework essentially combines the SplitFed, SplitFedv2/v3, and FedSL schemes in a hierarchical structure. Training on the clients in the same group works in the same way as in SplitFedv2 while the server-side models for different groups are trained in parallel and aggregated as in SplitFedv1. In addition, the server-side model training of different groups may be deployed either on a single server (like in SplitFed) or on multiple servers (as in FedSL). Therefore, SFLG is a generalized form of SplitFed (if one client per group and all groups are on a single server), SplitFedv2/v3 (if all clients are in one group served by a single server), and FedSL (if one client per group and one group per server). Therefore, SFLG offers a flexible architecture for combining federated and split learning with multiple possible configurations that can be chosen based on the learning objective as well as the available computing and networking resources in an IoT environment.

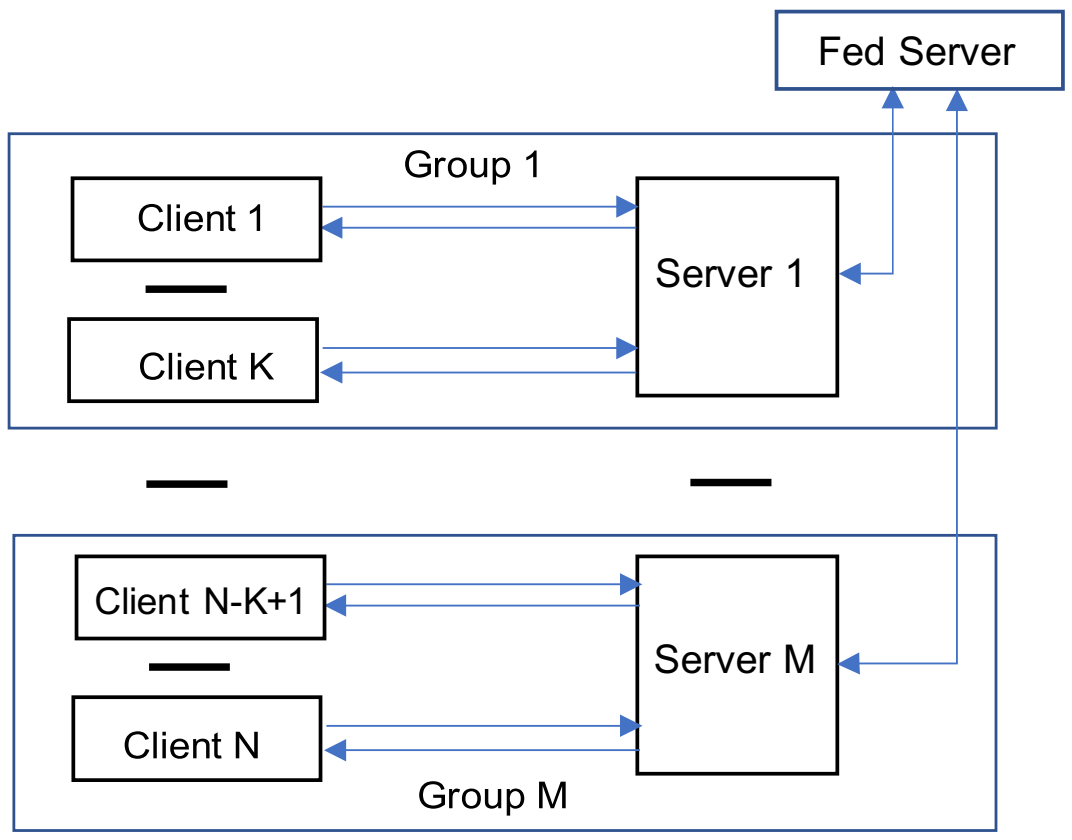

Fig. 5. Generalized SplitFed (SFLG) learning framework architecture.

The current representative proposals for hybrid FL-SL frameworks can be categorized from two aspects: i) the adoption strategy of federated parallel training – either on both the client and server sides or only on the client side, and ii) the architecture for server deployment – either centralized on a single edge node or distributed across multiple edge nodes. Table II gives a summary of the key features of the reviewed hybrid SL-FL frameworks from these two aspects.

Comparative study on learning performance and training costs of SL, FL, and hybrid SL-FL frameworks was conducted in [68] for various IoT settings. The learning performance of the frameworks was evaluated under iid, imbalanced, and non-iid data distributions with different number of clients. The obtained experiment results indicate that under iid and imbalanced data distributions SL typically converges faster than FL but suffers unstable convergence (shown by frequent spikes in the learning curve). It was also found that SL performance is affected by the number of clients which implies that FL may outperform SL even under iid data in an IoT setting with a large number of clients. Experiment results for non-iid datasets demonstrate that SL is much more sensitive to non-iid data compared to FL, which makes SL fails to learn in some cases with highly skewed data distributions while FL model still converges to a certain accuracy level. The learning performance of SplitFed (tested as a representative hybrid SL-FL framework) was found close to that of FL under all types of data distributions, which verifies the effectiveness of introducing federation in split learning for enhancing SL performance under non-iid data.

The SL, FL, and SplitFed frameworks were implemented on a Raspberry Pi platform for evaluating their computation and communication overheads in [68]. The results obtained from the experiment setting (training a small model on a limited number of clients) indicate that SL always has less computation overheads (including CPU/memory usage and power consumption) compared to FL but suffers a longer training time and more communication costs. It was found that SplitFed greatly reduces the training time with the same level of overheads as compared to SL, which demonstrates the benefit of combining SL with FL that achieves shorter training time than that of SL with less computation overheads on client devices compared to FL.

TABLE II
KEY FEATURES OF REPRESENTATIVE HYBRID SL-FL FRAMEWORKS.

| SL-FL Frameworks | Federated Parallel Training | Server Deployment |
|---|---|---|
| SplitFed [62] | on both client and server | centralized on a single server |
| SplitFedv2 [62] | only on client | centralized on a single server |
| SplitFedv3 [63] | parallel on client (alternative mini-batch on server) | centralized on a single server |
| CPSL [64] | parallel-first-then-sequential on client | centralized on a single server |
| HSFL [65] | mixed sequential SL and parallel FL | centralized on a single server |
| FedSL [66] | on both client and server | distributed to multi-servers |
| SFLG [68] | on both client and server | hierarchically distributed |

### B. Model Decomposition in Hybrid SL-FL Frameworks

A key aspect of SL and its combination with FL is to decompose an ML model and assign the resulted sub-models to clients and server(s). Proposals of hybrid SL-FL frameworks often focus on the collaboration among sub-models assuming a given cut layer; however, how the model is split has a significant impact on learning performance and training costs in split learning thus also deserving thorough investigation.

Some recent research on model decomposition considers the architectural features of specific ML models to determine the appropriate split of the models. For examples, in [69] the authors observed that the Vision Transformer, a recently developed deep learning model for image processing, has a design comprising three key components – a head for extracting input features, a transformer body for modelling the dependency between features, and a tail used for mapping features to task-specific outputs. Among the components, the transformer body is computing intensive and shared by various tasks. Based on this observation, the authors proposed a Federated Split Task-Agnostic (FESTA) framework in which multiple clients, each holds its own head and tail parts of the model, share a server hosting the transformer body. The server also aggregates the weights of local heads and tails from the clients via FedAvg to update the global head and tail, which are then distributed to all clients.

Another model specific decomposition approach is employed by the FedBERT framework proposed in [70]. FedBERT is a federated SL framework for pre-training the BERT model for natural language processing applications. The BERT model comprises three layers – an embedding layer, a transformer layer, and a head layer. The transformer layer is computing-intensive therefore should be trained on a server with sufficient computational resources, while the training of embedding and head layers can be deployed on resource-constrained client devices. In the FedBERT framework, each client performs forward propagation from the embedding layer to the transformer layer on the server. The server receives forward propagation outputs from all the clients and generates the output of transformer layer, which is then sent back to the client as input of the head layer. The client computes the final output of the head layer and calculates the gradients to start the back propagation along the reverse path. The authors proposed two training strategies for FedBERT – parallel training and sequential training. In parallel training, the server trains one transformer layer for each client in parallel and aggregates the transformers of all clients to a global transformer at the end of each training epoch. For sequential training, the server maintains a single transformer layer that is trained by different clients one by one using their own datasets.

The aforementioned methods for model decomposition leverage the design structures of specific models, which might not be applicable to general models. Also, these methods assume a static assignment of training workload between the clients and server, i.e., split the model at fixed cut layer(s); However, the available amounts of computing and networking resources on IoT devices may vary with time. Therefore, static split between the client- and server-side models lacks the flexibility to adapt to a dynamic IoT environment. As an attempt to address this challenge, FedAdapt was proposed in [71] as a hybrid FL-SL framework that is able to adaptively determine which portion of a model to be offloaded to a server based on the computational resources on the client devices and the network bandwidth between clients and the server.

FedAdapt employs the reinforcement learning (RL) technique to dynamically determine an offloading point (OP) for each client in each round of training. All the layers of the model that are behind the OP are offloaded to a server. In order to make FedAdapt scalable to a large number of user devices, a clustering-based method is employed to divide all clients into multiple groups according to their computational resources – all the client devices in the same group are assumed to be homogeneous in their computing capabilities. The RL agent determines the OP for each group, i.e., all clients in the same group split their models in the same way. The maximal local training time of the clients in a group is used as the input state for the RL agent. Output action of the RL for each group is a value in [0, 1], which is the percentage of the model workload that stays on the client and can be used directly to determine the OP (the cut layer for splitting the model). The RL agent uses the average time of each training round as the reward function to minimize the average training time of all devices. The experiment results obtained in [71] indicate that FedAdapt may achieve substantial reduction in average training and can adapt to changes in network bandwidth as well as heterogeneity in IoT devices. On the other hand,

FedAdapt introduces extra complexity and overheads caused by the RL agent and the clustering algorithm.

### C. Reducing Overheads of Hybrid SL-FL Frameworks

The aforementioned works on combining SL with FL mainly focus on collaboration among workers to enable federated parallel training in split learning without major change in the communication aspect of the SL framework. Therefore, the proposed hybrid SL-FL frameworks have basically the same level of communication overheads as the regular SL (or even more overheads with the additional data transfer for aggregating the client- and/or server-side models). Improving communication efficiency of hybrid FL-SL frameworks in an IoT environment thus becomes an important research topic.

One attempt to reduce the communication overheads of hybrid SL-FL frameworks is the Multi-Head Split Learning (MHSL) scheme proposed in [72]. MHSL essentially allows the clients in the SplitFedv2 framework to perform model training in parallel without synchronization through a federation server thus removing the communication and computation overheads associated with client-side model aggregation. Empirical study results demonstrated that performance degradation of MHSL compared to SplitFedv2 varies with datasets (1%-2% for MNIST dataset and about 10% for CIFAR-10 dataset). In addition, the experiment results in [72] were obtained using only iid data distributions. Therefore, whether acceptable performance can be guaranteed in SplitFedv2 without client-side synchronization still needs more thorough investigation.

In [73], the authors proposed a local-loss-based scheme called LoccalSplitFed to improve communication efficiency and shorten training latency of hybrid SL-FL. The proposed framework has the same architecture as SplitFed – parallel training and model aggregation on both client and server sides. Unlike SplitFed, the proposed framework introduces an auxiliary network on the client side, which is an extra layer that uses the cut layer outputs as the inputs to calculate a local loss function. Then the client performs back propagation to update the client-side model using the gradients obtained based on the local loss function. Therefore, the proposed framework only transmits activations from clients to server and sends no gradient from the server back to clients thus reducing the communication overheads roughly by half. In addition, the client-side model training does not need to wait the gradients to come back from the server therefore reducing the training latency. This may be particularly beneficial in an IoT environment with long communication delay due to limited bandwidth. The authors proved convergence for both client-side and server-side models and reported experiment results showing that the proposed framework outperforms SplitFed and FL in terms of both communication overheads and training time.

The techniques proposed in [72] and [73] focus on reducing the communication overheads for transmitting gradients from server back to clients and keep the transmission of smashed data from clients to server unchanged. However, the wireless communication systems in an IoT environment often have more constrained bandwidth on the up-links (from client devices to the server hosted on an edge node) compared to the down-links (from the server to client devices). Therefore, reducing the amount of smashed data transmissions from clients to the server is also critical to improve communication efficiency of hybrid SL-FL frameworks especially in an IoT environment.

The FedLite scheme proposed in [74] aims at reducing the up-link communication overheads in hybrid SL-FL while maintaining the accuracy of the learned model. The FedLite is based on an observation that, given a mini-batch of data, a client does not need to communicate per-example activation if the activations of different examples in the mini-batch exhibit enough similarity. Therefore, the FedLite framework performs clustering of the activations of each mini-batch of training data using an algorithm designed based on product quantization [75] and only communicates the cluster centroids to the server. Such activation clustering is equivalent to adding a vector quantization layer between the client- and server-side models, which may lead to drop in model accuracy due to the noised gradients that clients receive back from the server. In order to mitigate this issue, FedLite employs a gradient correction scheme that approximates the gradient by its first-order Taylor series expansion. The obtained empirical evaluation results show that FedLite is effective in achieving a high compression ratio with minimal accuracy loss. Although SplitFedv2 was assumed as the learning framework in [74], the proposed FedLite scheme is applicable to any hybrid SL-FL framework that can benefit from a vector quantization layer.

Binarization, as an extreme quantization method, has also been explored as a technique for reducing overheads in SL. Binarized Neural Network (BNN) are neural networks where weights and activations are constrained to either -1 or +1 for mathematical convenience [76]. Although BNNs consume less resources for computation and communications, they might not be able to achieve as good accuracy as their full-precision counterparts. In [77], the authors proposed to binarize the client-side models during forward propagation to speed up computation on clients and reduce communication overheads for transmitting activations. Meanwhile, the server-side model are kept in high-precision computation to retain accuracy of the global model. In the proposed binarized SL (B-SL) framework, the gradients are passed back to clients in full-precision during back propagation for updating the client-side models using a straight-through estimator [78]. Experiment results demonstrated that B-SL can reduce overheads while maintaining model accuracy; however, the effectiveness of B-SL for overhead reduction and its loss in model accuracy are directly impacted by how the model is split and still need to be thoroughly evaluated.

The existing approaches proposed for combining SL with FL typically employ techniques of parallel client-side model training and decoupling the training of client- and server-side models, which potentially suffer the problems of server-client update imbalance and client model detachment [79]. The server-client update imbalance occurs when the server-side model is trained with aggregated smashed data from multiple clients while the client-side models are trained only using the local datasets on individual clients. Such imbalance in model update, although can be mitigated by client-side model aggregation, may become an issue with a large number of clients thus limiting the scalability of hybrid SL-FL frameworks. Decoupling the client- and server-side model training, for example local loss-based client-side training in LocalSplitFed, prevents the client-side model update from utilizing the full capacity of a deep neural network therefore may cause the detachment of client-side model from the server-side model.

The LocFedMix-SL scheme was proposed in [79] for tackling the problems of imbalance between client and server side updates and detachment of client and server models. The operations of LocFedMix-SL partly coincide with those of LocalSplitFed [73] but employ some additional mechanisms. The key new elements are locally regularizing the sub-model at each client and augmenting smashed data at the server. The regularized local gradients maximize the mutual information between the raw and smashed data while avoid extracting too much of the original features on the client side. The server combinatorially superpositions smashed data uploaded from different clients to produce new augmented smashed data for the forward propagation on the server side to address the imbalanced update problem. In addition, considering the crucial role of the global gradients provided by the server to clients in avoiding model detachment, LocFedMix-SL resorts global gradient back propagation.

Although the LocFedMix-SL scheme offers an approach to addressing the update imbalance and model detachment issues associated with parallel SL, it introduces extra computation overheads (e.g., for smash data mix-up and local gradient regularization) as well as communication overheads. On the other hand, reducing communication overhead through removing client model synchronization as proposed in [72] may sacrifice learning performance due to loss of information sharing among clients. In order to achieve communication/computation efficient federated SL without sacrificing learning performance, the authors of [80] proposed a framework for split learning with gradient average and learning rate splitting (SGLR). In the SGLR framework, the update imbalance issue is addressed using an SplitLr algorithm that adjusts the learning rate for the server-side model according to the concatenated batch size on the server. For solving the model detachment problem SGLR employs an SplitAvg algorithm at the server, which calculates the weighted average of the gradients for a subset of clients and sends the averaged gradients to these clients for client-side model update. The experiment results reported in [80] show that the SGLR framework achieves comparable model accuracy as the baseline FL/SL and SplitFed algorithms but has less computation overhead (no smashed data mix-up) and communication overhead (no communication for client model aggregation) compared to the LocFedMix-SL framework.

### *D. Hybrid SL-FL Frameworks with Vertical and Sequential Configurations*

Although the majority of research on FL, SL, and their combination assumes a horizontal configuration, vertical FL recently started receiving more attention. Split learning offers a commonly used approach to addressing vertical federated learning especially for neural network models.

A vertical SL architecture proposed in [51] is illustrated in Fig. 6. In this architecture, each client $C_i$ holds a segment of features $X_i$ of the same set of data samples $X$. An ML model $F$ (e.g., a deep neural network) is split to the client- and server-side models, and the client-side model is further vertically split across the clients. That is, the partial model $F_i$ on client $C_i$ takes $X_i$ as the input. The outputs $S_i$ ($i = 1, 2, \cdots, K$) of all clients are transmitted to the server, which aggregates all client outputs to form an input $D = A(S_1, S_2, \cdots, S_k)$ to the server-side model $F_S$ to complete the forward propagation and calculates the loss function and gradients. During back propagation, the gradients $G_i$ for the respective $S_i$ are sent back to $C_i$ for updating the corresponding partial model $F_i$. In this architecture, the partial models on all clients are trained in parallel and aggregated together by the server who handles the training of the rest of the model; therefore, it essentially combines vertical SL and FL in the same architecture.

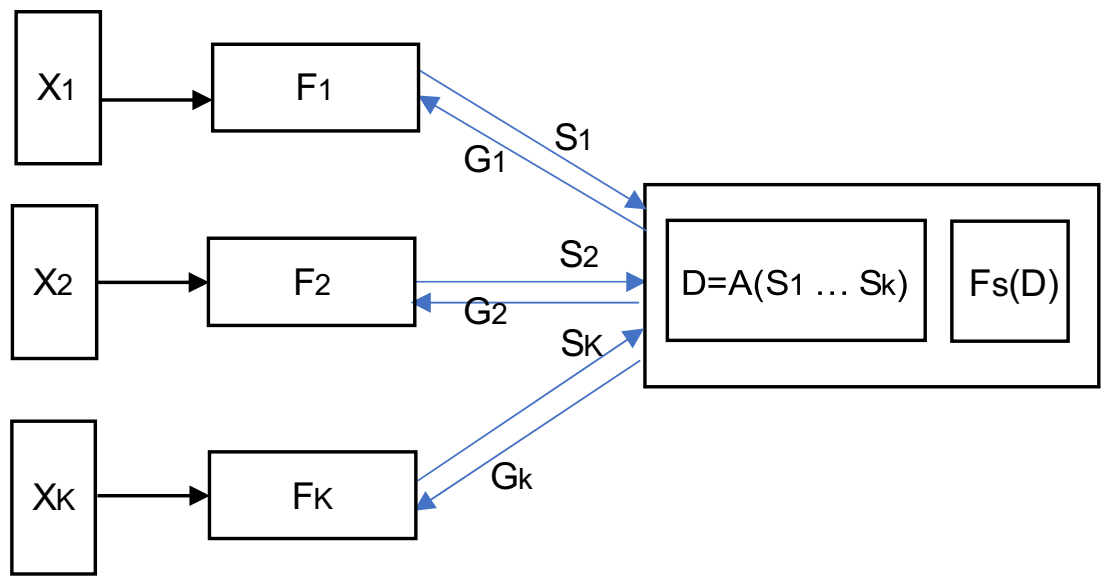

Fig. 6. Hybrid SL-FL framework with vertical configuration.

There are multiple mechanisms available for aggregating client outputs to form the input to the server-side model in a vertical SL-FL framework; for example, concatenation, element-wise average, element-wise sum, element-wise maximum, etc. Concatenation is a simple approach that is the closest to training a single model with all the input features. However, this method requires all clients to have intermediate outputs ready in each iteration thus is less robust to stragglers that are common in IoT where client devices have diverse computing and networking capabilities. The element-wise sum and element-wise average methods are similar to each other. Both methods require the partial models on all clients to have a compatible form so that their outputs can be combined together. On the other hand, these methods allow the use of a secure aggregation protocol for enhancing privacy and security of the learning framework. Element-wise maximum picks the activations with the maximum value for each neuron and discards the rest, which also requires the partial models on all clients to have compatible outputs.

The learning performance achieved by the vertical SL-FL framework using various aggregation mechanisms was evaluated in [51]. The obtained results indicate that no significant difference in the performance of these mechanisms, which are all close to that of centralized model training. The maximum pooling method appears to have the best overall performance while the average-pooling method is also very attractive due to its support for secure aggregation protocols.

Since different attribute values of the same data points are distributed across the clients in the vertical SL-FL configuration, it is critical to identify and link the data points on different clients that belong to the same data samples shared among clients. In SL/FL frameworks, such data alignment must be done in a way without exchanging either raw data or meta data among clients. Private Set Intersection (PSI), a multi-party computation cryptography technique, offers an approach to addressing this challenge. PSI allows two parties that each holds a set of elements to compute the intersection of these elements without revealing anything to the other party except for the elements in the intersection [81].

The PyVertical framework proposed in [82] employs the PSI technique for data alignment to support vertical FL using split neural networks. PyVertical uses a Diffie-Hellman key exchange-based PSI implementation with Bloom filters compression to reduce the communication complexity [83]. This PSI protocol works as follow. First, the server runs the PSI protocol independently with each data owner and the intersection of IDs between the server and each data owner is revealed. Then the server computes the global intersection from all the previously identified intersections and communicates the global intersection to the data owners. In this setting, the data owners do not communicate and are not aware of each other's identity thus facilitating client privacy protection.

Typical vertical FL algorithms assume that data features are split among multiple data owners while label information is held by a single entity (label owner). However, in some practical application scenarios there might be multiple label owners holding the labels of different samples. For example, multiple specialist hospitals (heart disease hospital, lung disease hospital, cancer hospital, etc.) have different feature data for the same set of patients whose COVID test results (the labels) are available at different testing centers. The Multi-Vertical Federated Learning (Multi-VFL) framework proposed in [84] aims to enable collaborative learning in such scenarios with multiple data owners and label owners.

The Multi-VFL framework makes use of split learning and adaptive federated optimization algorithms. In this framework, each data owner has a unique part of the client-side model and all label owners share the same server-side model. Data owners perform forward propagation on their respective partial models up to the cut layer and then send their activations to label owners. Label owners concatenate the activations coming from different data owners, complete their forward propagation, compute the loss, and perform back propagation. Label owners send the corresponding gradients back to data owners who will then use the respective gradients to complete their back propagation. The model parameters on label owners are aggregated by a Fed server using one of the FL algorithms such as FedAvg [9], FedAdam [85], FedYogi [85], etc. Adaptive federated optimization is employed in Multi-VFL for solving the model shift problem (local models move away from the optimal global model) caused by the non-iid data from different owners. PSI is also applied in this framework to identify intersections of the data sets vertically split across data owners.

In additional to horizontal and vertical data partition, sequential data partition also occurs in some ML-based IoT applications. One-dimension convolutional neural network (1DCNN) is an ML model that can be used for time-series classification and application of SL to 1DCNN with a single client was investigated in [52]. The most common ML models to train on sequential data are Recurrent Neural Networks (RNNs), including Long Short-Term Memory (LSTM) and Gated Recurrent Unit (GRU) [86]. However, existing SL approaches developed for feed-forward neural networks are not directly applicable to RNNs. In order to train on sequentially partitioned data, RNNs should be split in a way that preserves latent dependencies between the segments of sequential data available on different clients.

In order to address this issue, a framework that integrates SL and FL for training models on sequential data was proposed in [87]. In this framework, a RNN is split into sub-networks and each sub-network is trained on one client containing a single segment of the multi-segment training sequence. The sub-networks communicate with each other through their hidden states in order to preserve the temporal relation between the data segments available on different clients. Collaborative model training is performed across different clients following the SL pattern and the sub-networks trained on clients are aggregated by the server using an FL algorithm. An approach to split LSTM was proposed in [88], which considers another type of sequential data partition in which the entire length of time sequence input is stored on each client. The authors proposed to split LSTM from the $c$-th layer instead of splitting the network based on the input steps. Despite the recently reported progress on split RNN and LSTM, hybrid SL-FL with sequential data partition is still in its infancy and needs more research as we will discuss in Section VI.

## V. Privacy Protection in Split Learning

Similar to FL, SL is a collaborative learning paradigm that allows clients to jointly train a global model without sharing their own data, which protects privacy of users' data. Unlike FL, where the architecture and hyper-parameters of the global model are known to the server (for model aggregation), SL splits the entire model onto the client and server thus making the hyper-parameters of the global model unknown to the server, which may facilitate privacy protection. On the other hand, the intermediate representation (the smashed data) is revealed in SL, which actually introduces new vulnerabilities that attackers may exploit to compromise user privacy. A comparison of information exposure in SL and FL is given in Table III. Privacy and security of federated learning have been reviewed in a variety of surveys for example [89]. In this subsection, we focus on reviewing the unique privacy issues associated with SL and its combination with FL, and discuss the technologies developed for protecting privacy in SL/SL-FL frameworks.

### A. Privacy Attacks to Split Learning

*1) Attacks for Training Data Inference:* A pervasive vulnerability of SL lies in the server's entrusted ability to control

TABLE III
COMPARISON OF INFORMATION EXPOSURE IN FL, SL, AND HYBRID SL-FL.

| Information Revealed | Raw Data | Model Parameters | Intermediate Representation |
|---|---|---|---|
| FL | No | Yes | No |
| SL | No | No | Yes |
| hybrid SL-FL | No | Yes | Yes |

the learning process of clients' sub-models. Using such control ability, a malicious server is able to guide the client-side model toward functional states that can be exploited to recover the original training samples used on the clients.

In [90], the authors designed a general attack strategy Feature-Space Hijack Attack (FSHA) that allows a malicious server to recover private training samples during the training process. In FSHA, the server hijacks the learning process of client models and drives them to an insecure state that can be exploited for sample and property inference. An FSHA attack comprises two phases: a setup phase where the server hijacks the client-side learning process $f$, and a subsequent inference phase where the server can recover private training instances using smashed data sent from the client. FHSA requires the adversary to know a dataset $X_{pub}$ that captures the same domain of the client's training dataset $X_{priv}$. The attacker trains two components on the server – a discriminator $D$ and an auto-encoder $\tilde{f}/\tilde{f}^{-1}$. The auto-encoder, when given inputs from $X_{pub}$, encodes them using $\tilde{f}$ to an internal representation (feature space) and decodes using $\tilde{f}^{-1}$ with the loss set so as to learn to reconstruct the original input data. Finally, the discriminator $D$ is responsible for classifying the input as coming from either the $f$ model or the $\tilde{f}$ model. The attacker's target is for the discriminator $D$ to force the outputs of $f$ to come from the same feature space as those of $\tilde{f}$ and make $\tilde{f}^{-1}$ be able to decode $f$ with minimal error just as it learns to decode $\tilde{f}$.

FSHA equally applies to the hybrid SL-FL frameworks such as SplitFed variants since in these frameworks the server still maintains control of the learning process of client-side models. The only difference is in how the models are updated and synchronized among clients. Actually, FSHA may be potentially more effective in hybrid SL-FL frameworks as the server receives larger batches of smashed data that can be used to smooth the learning process of the discriminator. Once an attacker obtains a sufficiently good client model it can reconstruct the private training data using the smashed data passed from clients to the server during the training process.

Differential Privacy (DP) is a typical mechanism for privacy protection that has been employed in various federated learning frameworks. DP noise applied during model training is designed to prevent the model from memorizing data so that the private data cannot be retrieved from the trained model. However, it has been found in [91] that DP does not provide sufficient protection against FSHA. The authors of this paper applied FSHA to the learning process of an SL framework protected by DP using a client-side DP optimizer. The obtained experiment results indicate that although DP can delay convergence of FSHA, this attack method is still able to reconstruct the client's private data with a low error rate at an arbitrary DP setting. The author also explored applying dimensionality reduction techniques [92] directly to the raw data prior to training as a privacy protection mechanism and found that it may disable FSHA to some extent; however, it might also degrade accuracy of the trained model especially for learning tasks with large datasets.

*2) Attacks for Client Model Inversion:* Although FSHA may infer raw training data it is not able to steal the client-side model, which is often also an attacker's target. In addition, FSHA requires an attacker to have a public dataset following a similar distribution with that of the client's data. Unlike FSHA, the model inversion and stealing attack to SL designed in [93] aims to obtain a functionally similar model to the client model as well as the raw training data without the requirement of prior knowledge of the client's data set. The only assumption made by this attack method is that the client-side model architecture is known to the attacker. Without any data similar to the training data or the ability to query the client model, the attacker's task is essentially a search over the space consisting of all possible input values and client model parameters. This search can be modeled as an optimization problem in which the attacker tries to find model parameters $\tilde{\theta}$ and input $\tilde{x}$ that minimize the difference between $\tilde{f}(\tilde{\theta}, \tilde{x})$ and $f(\theta, x)$, where $\tilde{f}$ and $f$ are respectively the estimated and real client models. In [93], the authors employ the coordinate gradient descent approach to solve this optimization problem by alternatively fixing a subset of the parameters (e.g., $\tilde{\theta}$) and updating the rest of the parameters (e.g., $\tilde{x}$). Since the attacker strictly follows the SL protocol and only needs the intermediate activation values to conduct the model inversion and data inference, it is hard for clients to detect such an attack.

*3) Attacks for Private Label Leakage:* The training data and labels may be owned by two parties in some SL configurations; for example, the client and server are respectively data owner and label owner. Private label inference (leakage) attacks aim to enable the data owners (or any adversary) to infer private labels.

Li *et al.* proposed a norm-based label leakage attack to two-party split learning in [94]. This attack strategy exploits the class imbalance in the training datasets of some tasks (e.g., conversion prediction), which results in the gradient magnitude being higher when the infrequent class is encountered. Therefore, an adversary can infer the private class labels by analyzing the norm of gradients. However, this attack strategy is limited to binary classification problems with high class imbalance.

A gradient inversion attack scheme was devised in [95] for label inference in a more general setting of multi-class problems. In this scheme, the attack is formulated as a supervised learning problem by replacing the unknown parameters of the label owner with learnable surrogate variables. This allows an adversary to "replay" the split learning process with surrogate variables. The authors developed a loss function that matches the gradients collected during the replayed split learning with the gradients obtained during the original learning. The private labels can be recovered with high accuracy by minimizing this loss function. However, the gradient inversion attack requires the adversary to collect all the gradients and smashed data and have the full knowledge of the private label distribution, which may not be available in practical split learning settings.

A clustering label inference attack approach that is applicable to more practical SL settings was devised in [96]. The authors proposed the cosine and Euclidean similarity measurements for the gradients and smashed data at arbitrary layers. The label inference problem can be solved by the $K$-means clustering algorithm to classify the data points into $K$ clusters where $K$ is the number of classes of the classification task. This attack strategy does not require knowledge of the victim's model and is scalable in different SL settings (e.g., different positions of the cut layers) and robust to different defense techniques (e.g., differential privacy and gradient compression).

### B. Privacy Protection in Split Learning

*1) Techniques for Protecting SL against Data Inference and Model Inversion:* A key privacy vulnerability of SL lies in the fact that a neural network is a differentiable smooth function that is naturally predisposed to be functionally inverted. Splitting the neural network model to the client and server sides will not change this property. The intermediate results (the smashed data) exposed by an SL framework may be exploited to recover the original raw data and/or client model. Therefore, techniques for SL privacy protection against inference attacks typically aim at minimizing information leakage from the smashed data.

The method for protecting SL from inference attacks proposed in [97] reduces the information leakage by adding a distance correlation-based loss term to the loss function. Distance correlation is a measure of statistical dependence between random variables. The distance correlation loss is minimized between the raw data input and cut layer output. Optimizing the combination of these two loss terms (the regular loss function and the distance correlation loss term) helps ensure the activations from the cut layer to have minimal information with regards to reconstructing the raw data while still being useful enough to achieve reasonable model accuracy.

In [98], the authors proposed a method to defend against model inversion attacks. This method applies additive Laplacian noise to the cut layer activations before sending them to the server, which makes it harder for an attacker to learn the mapping from intermediate representation to input data. The authors conducted experiments to evaluate the effectiveness of the proposed noise-based defense method and compared it with the distance correlation-based defense proposed in [97]. The experiment results indicate that these two methods protect different information from exposure, which suggests that combining the distance correlation-based method with additive noise may produce a more robust defense.

The B-SL framework proposed in [77] applies binarization to the smashed data sent to the server, which contains latent noise added to the activations from the cut layer thus reducing the server's ability to reconstruct the raw training data. The authors also propose to exploit an additional loss term along with the model accuracy loss in order to minimize leakage of local private data. The loss term in B-SL is general that could be any leakage metric without being limited to the distance correlation term used in [97]. In addition, the authors provided three methods, namely double binarization, stochastic binarization, and randomizing response, to apply differential privacy in the B-SL framework for enhancing privacy protection. Experiment results reported in [77] demonstrated that B-SL is effective in mitigating the privacy vulnerability under FSHA attacks.

Application of the distance correlation-based protection method [97] to distributed ML settings such as hybrid FL-SL frameworks may result in a loss of privacy. This is because a distributed ML framework exposes too much information to be protected by a single distance correlation term in the loss function. For example, adding the correlation distance increment to the server's loss function could be a potential leak of information that enables an attacker to reconstruct the original input data if it has both the distance correlation value and the activations data sent over the network. Therefore, it would be desirable to force clients and servers to use different loss functions to avoid such privacy leakage.

Toward this direction, the authors of [67] proposed a client-based privacy protection approach in a hybrid FL-SL framework. This approach uses two different loss functions computed on the client and server sides respectively in order to overcome the limitation caused by a single global loss function in a distributed ML setting. The first loss function is privacy-aware (e.g., distance correlation or differential privacy) and runs only on clients. The second global loss function is computed on the server and propagates across both clients and server during the training process. The obtained experiment results show that the proposed approach can keep data privacy in both hybrid FL-SL and parallel SL frameworks. It was also found that the client-based privacy approach using distance correlation achieves better performance in terms of privacy and model accuracy compared to the DP-based approach.

In order to defend hybrid SL-FL frameworks from model inversion (MI), Li *et al.* [99] designed an MI-resistant framework ResSFL that comprises two steps – a pre-training step that builds up a feature extractor with strong MI resistance and a follow-up resistance transfer step that initializes the client-side models using the feature extractor. The pre-training step employs the attacker-aware training technique to emulate an attacker using a strong inversion model and adds bottleneck layers to the inversion model to reduce the feature space. This step is implemented on an "expert" (i.e., one powerful client or a third party) that has sufficient computation resources for performing the pre-training on a source task. Then in the second step, the resistant feature extractor is used as an initialization for the SL-FL training scheme on a new task.

Splitting the model training to the client and server sides in SL introduces a unique type of inference attack that allows a malicious server to hijack the training of client model thus inferring the training data, as FSHA proposed in [90]. In order to deal with this type of attacks, the authors of [100] proposed SplitGuard, a method by which an SL client can detect whether it is being targeted by a training-hijack attack. The intuition behind the proposed SplitGuard method is an observation that if the training process of client model is being hijacked then the process will behave in a drastically different way than what it should do for learning the intended task.

During training with SplitGuard, clients intermittently input fake batches - batches with randomized labels. The main idea of SplitGuard is that if the client model is learning its intended task then the gradient values received from the server should be noticeably different for the fake batches and regular batches. More specifically, the gradients for fake batches should have a higher magnitude than the regular gradients and the angle between the fake and regular gradients will be higher than the angle between two random subsets of regular gradients. Each client in SplitGuard keeps computing a score based on the fake and regular gradients that it has collected up to that point and decides whether the server is launching a training-hijack attack based on the history of the score values. The experiment results reported in [100] demonstrate that when the proposed SplitGuard enables SL clients to successfully detect training-hijack attacks when it is used appropriately in a secure setting without label-sharing.

*2) Mechanisms for Protecting SL against Label Leakage:* Label leakage is another type of privacy attack that is unique in SL frameworks where the data and labels are owned by different parties. In order to protect SL against label leakage in such a setting, the label owner (e.g., the server) should ideally deliver essential information about gradients without communicating the actual values. Random perturbation methods generally aim to achieve this goal thus are employed in the recent proposals for defending against label leakage in SL.

In [94], the authors proposed several random perturbation techniques to limit the label-stealing ability of the non-label party in a two-party SL framework. Among them, the main method "Marvell" (optiMized perturbAtion to pReVEnt Label Leakage) directly searches for the optimal structure of random perturbation noise to minimize the label leakage against a worst-case adversarial non-label party.

TPSL (Transcript Private Split Learning) proposed in [101] is a generic gradient perturbation-based SL framework that focuses on protecting label information using DP. TPSL leverages a gradient-perturbation scheme GradPerturb that adds noise only on the optimal direction of the gradients sent from the server (the label owner) to the client (the data owner). Experiment results obtained using large-scale datasets show that TPSL with Laplace perturbation can achieve good trade-off between model accuracy and label privacy.

In order to tackle privacy leakage of both data and label information, Xiao *et al.* [102] proposed a Multiple Activations and Labels Mix (MALM) mechanism that can not only protect raw data from reconstruction attacks but also obfuscate the label of each training sample. MALM makes use of sample diversity to mix the activation outputs from client-side models and create obfuscated labels before sending them from clients to the server in an SL framework. The mixed activations preserve a low distance correlation with raw data thus preventing private data from being reconstructed and the mixed labels can obfuscate the adversary thus avoiding leakage of the ground-truth label information.

## VI. Directions for Future Research

Although encouraging progress in split learning and its combination with federated learning has been made toward enabling ubiquitous intelligence in IoT, research in this area is still on an early stage with many open issues that still need to be thoroughly investigated. In this section, we attempt to identify some open problems for further study and discuss possible directions for future research in this emerging field.

### *A. Model Decomposition and Deployment for Hybrid SL-FL*

Most of the current SL and hybrid SL-FL techniques assume static model split between the client and server, which lacks the flexibility to adapt to the heterogeneous and dynamic IoT environment. Although early effort has been made in [56] toward adaptive model split, this proposal only considers the resource availability on client devices and assumes sufficient capacity on the server. However, in an edge computing-based IoT environment various edge nodes with highly diverse computing and networking capabilities may be used for hosting an SL/FL server. Therefore, the resource availability on edge nodes should also be fully considered in the process of model decomposition. In addition to model decomposition, deploying the offloaded model portion on the most appropriate edge node(s) is also crucial to achieve satisfactory learning performance while fully utilizing resources of the edge infrastructure. Existing SL and hybrid SL-FL techniques split an ML model to two portions (for the client and server) without exploiting the flexibility offered by the extended SL configuration, in which the server-side model may be further split to multiple portions that are deployed on different servers. Such a configuration enables flexible learning frameworks utilizing the edge computing platform but calls for more sophisticated techniques for choosing the best edge nodes for hosting all the servers.

Therefore, the problem of model decomposition and deployment in edge computing deserves more thorough investigation in future research on combining SL and FL. The main questions to be answered include which portion of the model should be offloaded from a user device and where the offloaded portion of the model should be deployed (either on a single server or across multiple servers). Multiple factors, including learning performance, training costs (computation and communication overheads), and privacy/security vulnerabilities, should be jointly considered when searching for an optimal solution. Fully examining the architectural features of ML models might provide useful insights and guidelines. Recent research on model splitting in the context of distributed inference, for example the work reported in [103] and [104], also offers ideas that could be borrowed for tackling model decomposition/deployment from the training perspective. Model decomposition and deployment in SL-FL combination share essential similarity with the task offloading and scheduling problems that have been extensively studied in the field of edge computing. Various technologies have been proposed for computational offloading and scheduling in edge computing [105]; however, little work on their application to SL/hybrid SL-FL has been reported in the literature. Therefore, exploring the rich results about offloading and scheduling from edge computing research to tackle the challenge of model decomposition and deployment in combining SL and FL offers a promising direction for future research.

### *B. Cross-Layer Resource Management for Hybrid SL-FL*

The recent progress on hybrid SL-FL techniques introduces learning frameworks with servers for performing different tasks, including synchronizing the server- and client-side models as well as training the offloaded portion of models. The tasks executed on these severs require different amounts of computational and communication resources. For example, a server hosting the parallel training tasks for multiple server-side models consumes large CPU capacity and memory space for data processing and storage while a server performing model aggregation for a large number of clients requires high bandwidth for data transmissions. In addition, the flexible configuration of hybrid SL-FL frameworks, for example the hierarchical structure of SFLG proposed in [68], enables various communication patterns among different servers. On the other hand, recent development in edge computing technologies enable various flexible edge-edge and edge-cloud collaborations in IoT [106]. Therefore, flexible and efficient resource management for supporting hybrid SL-FL frameworks upon an edge computing platform becomes a challenging research problem that needs thorough study in the future.

The hybrid SL-FL frameworks can be regarded as a special type of applications deployed upon the platform provided by the edge computing infrastructure. Therefore, resource management in the underlying edge computing system for supporting the hybrid SL-FL applications calls for a holistic vision across the application layer and the infrastructure layer. A key aspect of the resource management lies in optimal mapping from application requirements (achieving an acceptable level of learning performance) to infrastructure resource allocation (server deployment and network connections). Hybrid SL-FL involves both computation (e.g., for model training) and communication (e.g., for exchanging activations and gradients) functions while the edge-cloud infrastructure comprises both computing and networking systems. Therefore, another aspect of the holistic resource management for hybrid SL-FL is about end-to-end allocation of computation and communication capabilities in the edge-network-cloud continuum. Virtualization and service-oriented architecture are two key pillars for edge-network-cloud convergence [107], which enables a wide variety of upper-layer applications to share a unified computing-networking infrastructure. Further investigation on virtualization-based service-oriented resource management in edge computing for supporting SL-FL combination in an IoT environment offers an interesting direction for future research.

### *C. Communication Efficient Hybrid SL-FL*

A key aspect of SL and its combination with FL lies in the trade-off between computation and communication in a learning framework. Most of the existing works on SL and hybrid SL-FL techniques tend to focus more on improving computation efficiency, which allows the learning frameworks to be flexibly deployed upon a resource constrained edge computing platform in an IoT environment. However, such computational efficiency and deployment flexibility often come with a higher communication cost. Communications play a critical role in hybrid SL-FL not only for exchanging intermediate data (e.g., activations and gradients) between split models but also for transmitting model parameters for aggregation. Frameworks with multiple servers deployed on different edge nodes rely on inter-server cooperation and generate more communication overheads. On the other hand, the network connections in an edge computing-based IoT are often implemented based on wireless mobile communication systems that provide constrained and inconstant bandwidth. Therefore, communications may become a performance bottleneck for hybrid SL-FL in IoT and communication efficient hybrid SL-FL deserves more thorough investigation in future research.

In general, the research problem of communication efficient hybrid SL-FL may be tackled from two aspects: i) reducing communication overheads in SL-FL frameworks, and ii) improving network resource utilization in the edge infrastructure. A variety of technical strategies could be explored for reducing SL-FL communication overheads; for example, reducing the frequency of data transmissions between the clients and server(s) through balancing local and global training, reducing the amount of data exchanged between the clients and server using quantization and sparsification techniques, and reducing the number of communication connections by employing selective update of client-side models. However, technologies for reducing communication overheads in SL-FL frameworks often sacrifice learning performance and introduce extra computing complexity. Therefore, how to achieve an optimal balance among communication efficiency, computation complexity, and learning performance is a research topic that particularly needs more study in the future. Regarding the aspect of improving network utilization for supporting hybrid SL-FL, research on resource-aware model decomposition/deployment and cross-layer management of computing-networking resources as discussed above may offer promising approaches toward communication efficient hybrid SL-FL.

### *D. Alternative Architectures for Hybrid SL-FL Frameworks*

Most of the existing frameworks for combining SL-FL have a centralized architecture in which one server works with a group of clients for collaborative learning. Although the recently proposed FedSL [60] and SFLG [68] frameworks distribute model training to multiple servers, model aggregation is still centralized on a single server. Such a centralized architecture faces some challenges in a large-scale dynamic IoT environment. A central server may become a performance bottleneck and a single point of failure when handling model training and/or aggregation for a large number of clients.

Data transmissions from a large group of clients to a central server (for collaborative training and/or model synchronization) likely cause network congestion especially in IoT with constrained network bandwidth. In order to addressing these challenges for further enhancing SL-FL combination in an IoT environment, exploring alternative architecture, especially decentralized architecture, for hybrid SL-FL frameworks offers a research direction that deserves thorough investigation in the future.

There are a variety of architectural options that could be explored for hybrid SL-FL frameworks. Representative options for decentralized architecture include hierarchical architecture where servers are organized into different levels (typically for handling local and global model training/synchronization), peer-to-peer architecture in which all servers are on the same level and cooperate with each other directly, and hybrid architecture that combines the hierarchical and peer-to-peer options. Decentralized FL recently attracted increasing research interest and various proposals have been made for FL frameworks with hierarchical, peer-to-peer, and hybrid architecture, including applications of Gossip protocol [108] and blockchain technologies in FL [109]. How to exploit the research results about decentralized FL to enhance hybrid SL-FL in an IoT environment would be an interesting topic for future research. Also, the problems of model decomposition/deployment, resource management, and communication efficiency in decentralized hybrid SL-FL frameworks all need to be fully studied in the future.

### *E. Performance Evaluation of Hybrid SL-FL in Edge-based IoT*

With more frameworks for hybrid SL-FL are proposed, it is important to obtain deep insights about various aspects of the performance, including convergence rate, model accuracy, and associated computation and communication overheads, of the proposed schemes and identify the key factors that impact the learning performance and training costs of the frameworks. The existing proposals of hybrid SL-FL schemes are typically supported by experiment results of performance evaluation obtained with their individual experiment settings, which makes comparison among the achievable performance of different frameworks not straightforward. Although some works on comparative study have been reported in the literature, they are either limited to the basic SL structure (as in [50]) or assume a relatively simple IoT scenario where a hybrid SL-FL framework is deployed (as in [68]).

Therefore, thorough evaluation on the performance that can be achieved by various hybrid SL-FL frameworks in an IoT environment is another topic that deserves more research in the future. It is critical to come up with general methods (e.g., a standard benchmark) for comparative performance study for both existing and emerging SL-FL frameworks in various deployment settings, which plays a key role in choosing the most appropriate learning framework for a given environment where the learning tasks are expected to be deployed. Performance evaluation for hybrid SL-FL upon an edge computing platform in practical IoT settings considering the unique features such as mobility of both user devices and edge nodes is particularly needed thus offering an interesting topic for future research. In addition to experiment-based empirical evaluation of learning performance and training costs, if analytical modeling for hybrid SL-FL frameworks can be exploited for obtaining useful insights about SL-FL performance is also worth more investigation.

### *F. Privacy and Security of Hybrid SL-FL*

One of the main objectives of SL and FL is to protect user privacy and data security through collaborative learning using local data without exposing training samples; therefore, privacy and security form an important aspect of SL-FL combination. Although privacy and security of FL have been extensively studied, the currently available research on this aspect of SL mainly focuses on privacy – possible attacks and protection mechanisms for private training data, client model, and label information. Little work dedicated to the security aspect of SL has been reported in the literature. SL security is about protecting the integrity of learning process to ensure accuracy and completeness of the trained models. Recent research has shown that malicious participants can influence or even control the model training process in collaborative learning for compromising model integrity [110]. However, how such compromise may be achieved and defended in an SL framework has not been clearly understood thus demanding more thorough study in future research.

Moreover, little research has been conducted on the privacy and security aspect of SL-FL combination. In a learning framework that combines split learning with federated model aggregation, both intermediate representation (activation data and gradient values) and model parameters are exposed; therefore, hybrid SL-FL introduces extra vulnerabilities that may be exploited by an adversary for compromising data privacy and/or model integrity. Therefore, analysis on possible privacy and security vulnerabilities and development of defense mechanisms for hybrid SL-FL frameworks form an important problem that is open for future research.

### *G. Combining SL and FL with Vertical and Sequential Configurations*

Although vertical FL with data feature partition recently started attracting more attention, the majority of the existing research on hybrid SL-FL still focuses on the horizontal configurations of both SL and FL and much less amount of work considers combining SL and vertical FL. Interesting progress toward this direction was recently published in the literature, for example in [51], [82], and [13]; however, a variety of aspects of vertical SL-FL combination are yet to be fully investigated. Some open issues for further study include design and deployment of vertical SL-FL frameworks that match the underlying edge platform, analysis and selection of the mechanisms for aggregating vertically split client models, evaluation of performance, costs, and scalability of vertical SL-FL in an IoT environment, and privacy and security issues related to vertical SL-FL frameworks. Therefore, combining SL and FL with the vertical configuration in order to further unleash the advantages of these two complementary learning methods deserves more thorough study in future research.

The current research on SL and its combination with FL often considers deep neural networks (e.g., CNNs) as the assumed models; therefore, the techniques developed for such feed-forward neural networks are not directly applicable to other popular ML models including sequential models such as RNN, LTSM, and GRU. However, ML for processing sequentially partitioned data is also needed in some IoT application scenarios. Although early efforts for extending hybrid SL-FL to sequential models have been reported, for example in [87] and [88], performance of the proposed frameworks in various IoT settings is yet to be fully evaluated and the extensibility of the proposed methods to other sequential models also needs to be further studied. Since research in this area is still on an infant stage, thorough investigation on hybrid SL-FL techniques for more general sequential ML models offers another important topic for future research.

## VII. Conclusions

Federated learning and split learning are two emerging collaborative learning methods that may complement each other toward enabling ubiquitous intelligence in IoT. In this article, we first review the latest developments of federated learning and split learning in the context of edge computing-based IoT. Then we present a survey on the state-of-the-art technologies for combining split learning with federated learning for facilitating ubiquitous intelligence in IoT. The survey indicates that encouraging progress has been made in various aspects of SL-FL combination, including hybrid SL-FL frameworks, model decomposition methods, techniques for enhancing performance and efficiency, and privacy protection. On the other hand, research in this area is still on an early stage and many open issues are to be further studied. In this article, we also identify some open problems and discuss possible directions for future research in this emerging field.